\documentclass[]{nature_mod}
\usepackage{amsmath,amsfonts}

\usepackage{txfonts}
\usepackage[english]{}
\usepackage[round]{natbib}
\usepackage[normalem]{ulem}
\usepackage{xr}
\usepackage{graphicx}
\usepackage{tocloft}
\usepackage{rotating}
\usepackage{booktabs}
\usepackage[table]{xcolor}
\usepackage{multirow}
\usepackage{adjustbox}
\usepackage{verbatim}
\usepackage{amsmath,amssymb}
\usepackage{algorithm}
\usepackage{algpseudocode}
\usepackage[hidelinks]{hyperref}

\definecolor{MyRed2}{rgb}{0.7,0.2,0.2}
\definecolor{MyBlue}{rgb}{0.2,0.2,0.7}
\definecolor{MyGreen}{rgb}{0.2,0.7,0.2}

\definecolor{tableShade}{HTML}{F1F5FA} 
\definecolor{MyGray}{rgb}{0.95,0.95,0.95}

\makeatletter 
\newcount\SOUL@minus
\makeatother  

\title{A Deep Generative Framework for Joint Households and Individuals Population Synthesis}

\author{Xiao Qian$^{1}$, Utkarsh Gangwal$^{2}$, Shangjia Dong$^{3}$ \& Rachel Davidson$^{4}$}

\begin{document}

\maketitle

\begin{affiliations}
 \item Graduate Research Assistant, Department of Civil and Environmental Engineering, University of Delaware, Newark, DE 19716. USA.
 \item Graduate Research Assistant, Department of Civil and Environmental Engineering, University of Delaware, Newark, DE 19716. USA.
 \item Corresponding author, Assistant Professor, Department of Civil and Environmental Engineering, University of Delaware, Newark, DE 19716. USA (sjdong@udel.edu)
 \item Professor, Department of Civil and Environmental Engineering, University of Delaware, Newark, DE 19716. USA.
\end{affiliations}

\begin{abstract}
\textbf{Abstract} Household and individual-level sociodemographic data are essential for understanding human-infrastructure interaction and policymaking. However, the Public Use Microdata Sample (PUMS) offers only a sample at the state level, while census tract data only provides the marginal distributions of variables without correlations. Therefore, we need an accurate synthetic population dataset that maintains consistent variable correlations observed in microdata, preserves household-individual and individual-individual relationships, adheres to state-level statistics, and accurately represents the geographic distribution of the population. We propose a deep generative framework leveraging the variational autoencoder (VAE) to generate a synthetic population with the aforementioned features. The methodological contributions include (1) a new data structure for capturing household-individual and individual-individual relationships, (2) a transfer learning process with pre-training and fine-tuning steps to generate households and individuals whose aggregated distributions align with the census tract marginal distribution, and (3) decoupled binary cross-entropy (D-BCE) loss function enabling distribution shift and out-of-sample records generation. Model results for an application in Delaware, USA demonstrate the ability to ensure the realism of generated household-individual records and accurately describe population statistics at the census tract level compared to existing methods. Furthermore, testing in North Carolina, USA yielded promising results, supporting the transferability of our method. 
\end{abstract}

\section{Introduction}\label{sec:intro}

Urban planning \citep{maantay2007mapping, sodiq2019towards, zhu2014synthetic}, disaster response and emergency management \citep{birkmann2006measuring, he2016population}, household adaptation behaviors analysis \citep{soleimani2023household}, and healthcare planning \citep{bouttell2018synthetic, gangwal2023critical} can all benefit from an accurate population dataset. With ever-increasing attention to equity and environmental justice in decision-making, there is a heightened imperative to conduct household-level investigations to capture the heterogeneous behaviors within the community \citep{chen2021unobserved}. Central to this effort is a comprehensive and accurate population dataset, serving as the cornerstone for the analysis and mapping of interactions between humans, the built environment, and external factors like disaster disruptions and policy interventions. However, due to privacy and other concerns \citep{cbs2022, globa2023Legal}, access to a complete true population dataset is often restricted and only anonymized samples and aggregated totals are available. The lack of a population dataset hampers a nuanced understanding of interactions between humans and the built environment at large. Therefore, there is a pressing need to create a realistic synthetic population dataset. In this work, we focus on joint household-individual population datasets in which each individual is defined by their values on a set of individual attribute variables (e.g., gender, age), and each household is comprised of one or more of those individuals and is similarly defined by its values on a set of household attribute variables (e.g., household income). 

Ideally, a synthetic population dataset should possess the following key features: \vspace{-6pt} 
\begin{enumerate}
    \item Each individual is realistic. That means each individual's characteristics should match real-world correlations. For example, there should not be a lot of very high-income 18-year-olds or teenagers who have doctoral degrees. \vspace{-6pt} 
    \item Each household is realistic. Namely, correlations among household variables should mirror those found in actual households. Additionally, the relationships among individuals within a household should align with real-world patterns. For example, households with individuals holding advanced degrees are more likely to have higher incomes, while lower-income families typically possess fewer vehicles. \vspace{-6pt} 
    \item The overall population is realistic. The synthetic population's marginal distributions of individual and household variables should match those observed in real populations at the state level. For example, the synthetic population should reflect the correct proportion of wealthy households as indicated by state statistics. This ensures that the synthetic population accurately represents the characteristics of the actual population. \vspace{-6pt} 
    \item The geographic distribution of population is realistic. Because population characteristics in different regions vary significantly, the marginal distributions of individual and household variables in the synthetic population should correspond to the ground truth marginal distributions. For example, the distribution of high-income households should match the wealth pattern in real life.
\end{enumerate}

\textbf{Data Challenge} Extensive efforts have been made to create synthetic population datasets, some even incorporated aspects of the housing unit characteristics \citep{rosenheim2021integration} or workplace assignment \citep{fournier2021integrated}. Public data sources such as the American Community Survey (ACS) and American Housing Survey (AHS) are commonly used for synthetic population development. However, the varying scales of the available population data samples and distributions make synthetic population generation a unique challenge. 

The ACS Public Use Microdata Sample (PUMS), hereafter referred to as \textit{microdata}, is released annually by the United States Census Bureau and offers detailed records of individual people and housing units \citep{PUMS2022}. These records cover a wide range of social, economic, housing, and demographic characteristics. With multiple variables for each individual and household, they provide the correlations among variables. Unfortunately, ACS PUMS is state level and based on a sample of 1\% for a single year or 5\% for five years. Researchers may also wish to deploy surveys to collect additional attributes that are not captured by microdata. In such cases, using private data to generate a synthetic population must ensure the privacy of the original human-subject data is preserved.

\begin{figure}[!ht]
    \centering
    \includegraphics[scale=0.55]{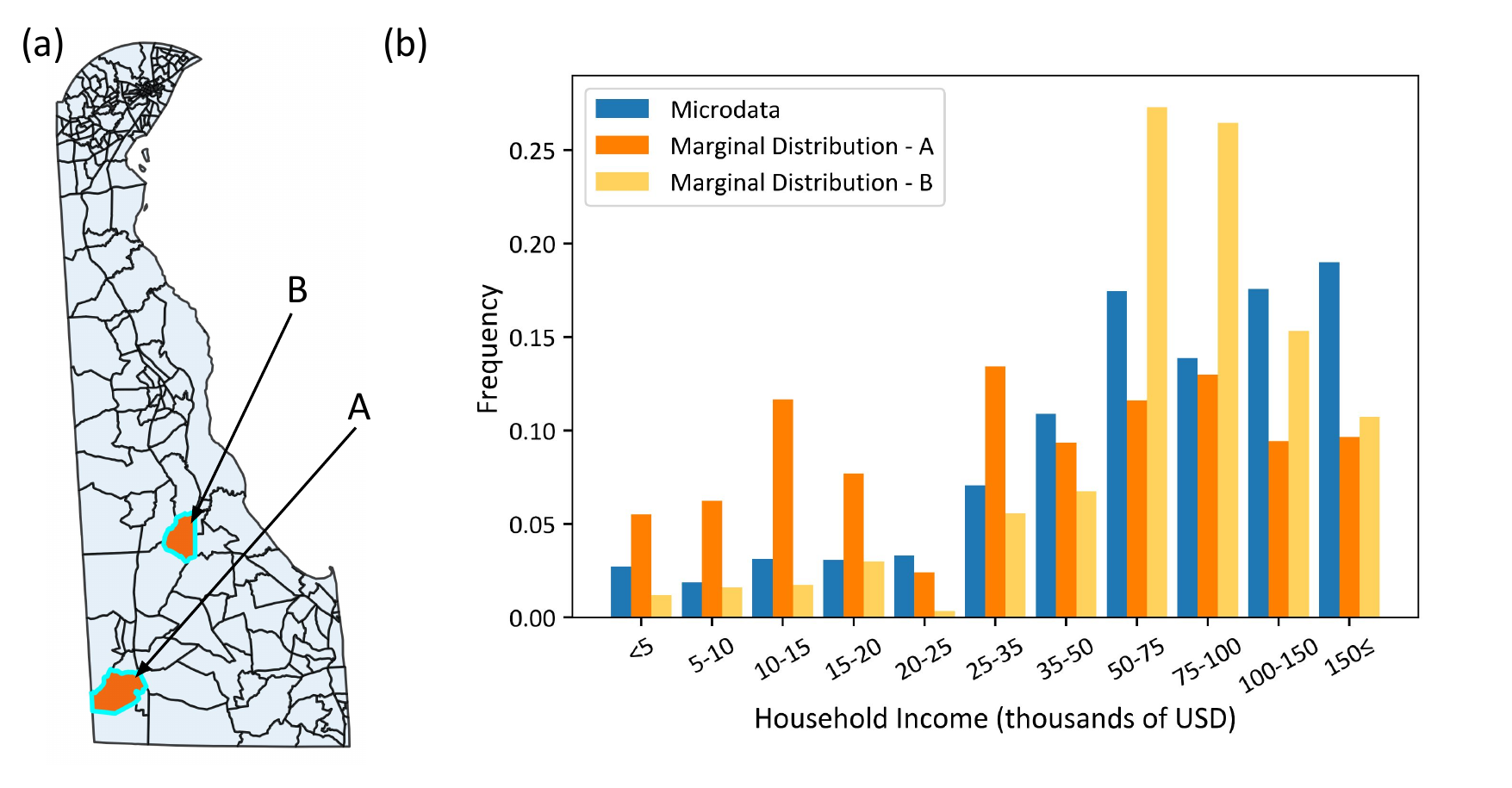}
    \caption{Comparison of household income distribution. (a) Map of the state of Delaware with two randomly selected census tracts, A and B; and (b) marginal distributions of household income for the state (microdata), census tract A, and census tract B.}
    \label{fig:data-challenge}
\end{figure}

The ACS also provides data at the census tract or block group level, hereafter referred to as \textit{census tables}, but it includes only marginal distributions of selected attributes \citep{ACS2022}. Due to the geographic distribution of the population, the marginal distribution of each variable may differ across census tracts and between the census tracts and state-level marginal distributions in the microdata (e.g., Figure \ref{fig:data-challenge}). Ideally, the individual and household variables that describe the synthetic population should exhibit the correlations from the microdata and the marginal distributions for each census tract from the local data.

\textbf{Research Gap} Extensive efforts have been devoted to developing synthetic populations, with optimization-based methods like Iterative Proportional Fitting (IPF) \citep{beckman1996creating} and Gibbs sampling \citep{Farooq2013Simulation}, or their derivatives \citep{ye2009methodology}, being widely utilized. However, they often suffer from what is known as the curse of dimensionality, where their effectiveness diminishes drastically with an increase in the number of attributes during synthetic population generation. Furthermore, they are limited to replicating existing samples rather than conducting true synthesis, losing the heterogeneity that was not captured in the microdata \citep{Farooq2013Simulation}. Deep generative approaches, including Variational Autoencoders (VAE) \citep{borysov2019generate, aemmer2022generative}, Generative Adversarial Networks (GAN) \citep{zhao2021ctab, zhao2022ctab}, and Diffusion Models \citep{kotelnikov2023tabddpm, lee2023codi}, offer solutions by generating out-of-sample data with numerous attributes. Nonetheless, these methods often fall short of generating population datasets that conform to the tract-level target marginal distribution. Existing deep learning methods, in particular, can only produce synthetic populations whose distribution aligns with the joint distribution of the microdata on which they are trained. Because the training data (i.e., microdata) distribution does not align with the target marginal distribution at the census tract level, as shown in Figure \ref{fig:data-challenge}, models that are trained, validated, and tested on microdata (only available at the state level) cannot accurately depict the population landscape at the census tract level. 

\textbf{Contributions} In this research, we introduce a novel deep-learning population synthesis framework with both household and individual characteristics embedded, aiming to include all key features outlined for an ideal synthetic population dataset. The primary technical contributions of this work can be summarized as follows: 
\begin{itemize}
    \vspace{-4pt}
    \item We propose a table restructuring technique to facilitate the learning of households-individuals and individuals-individuals relationships in microdata (Feature 2), enabling the generation of synthetic household and individual inventory simultaneously. This data representation streamlines the learning and generation process, overcoming the limitations of the conventional two-step approach of first generating synthetic individuals and then assembling them into households, which fails to capture the relationships between individuals who live in the same household. \vspace{-6pt}   
    \item We present a novel parameter-efficient transfer learning algorithm, that enables the adaptation of generative models trained on state-level microdata to produce synthetic households and individuals at the census tract level while conforming to the target marginal distributions from the ACS census data table (Features 3 \& 4). This method preserves the realism of individual household and individual records as depicted in microdata. Beyond population synthesis, the proposed algorithm can also be applied to other learning and generative tasks, particularly those with differing distributions between training and target data.\vspace{-6pt}
    \item We introduce a new loss function, Decoupled Binary Cross-Entropy (D-BCE), aimed at gauging the realism of synthetic data by quantifying the difference between the synthetic data and real samples (i.e., microdata) (Features 1 \& 4). 
\end{itemize}

\section{Related Works}\label{sec:lit-review}

Existing literature to generate the synthetic population with both households and individuals can be grouped into four main categories: (i) synthetic reconstruction (SR), (ii) combinatorial optimization (CO), (iii) statistical learning (SL), (iv) deep generative methods \citep{fabrice2021comparing, sun2018hierarchical}.
\subsection{Synthetic reconstruction}
Methods in this category typically follow a two-step process: fitting (where non-integer weights are assigned to individuals and households to match the marginal totals) and allocation (where these non-integer weights are converted to integers and individuals are then replicated based on these weights accordingly). A widely used SR technique is Iterative Proportional Fitting (IPF), which involves building a contingency table to match the marginal totals by minimizing discrimination information or relative entropy \citep{pritchard2012advances, little1991models}. While the IPF is simple and fast, its original formulation is incapable of generating both household characteristics and individual attributes concurrently. Researchers trying to use IPF to create a joint distribution of households and individuals either fit the household and individual attributes separately or sequentially, resulting in inconsistent fitting \citep{arentze2007creating, zhu2014synthetic}. To overcome this limitation, researchers have turned to two-layered population generation methods such as hierarchical iterative proportional fitting \citep{muller2011hierarchical, muller2017generalized} and iterative proportional update (IPU) \citep{balakrishnaa2019enhanced}, which group individuals into households while satisfying marginal totals at both levels. Hierarchical IPF or IPU entails iteratively computing weights for individual and household records, with cross-categorization of individual types into different household types \citep{chapuis2019brief, jain2015creating}. Synthetic population generation can also be viewed as a constrained optimization problem. The goal of optimization model formulations is to calculate household weights so that the weighted distribution of various attributes aligns with population distributions. One commonly used optimization model is entropy maximization \citep{barthelemy2013synthetic, paul2018multi, wu2018generating}. This approach aims to generate a synthetic population that closely aligns with specified marginal distributions by maximizing entropy while adhering to constraints derived from the sample population data \citep{lee2011cross}. By maximizing entropy, this model introduces diversity and randomness into the synthetic population, effectively safeguarding the privacy of sample population data. Researchers have also explored other optimization-based models such as generalized raking \citep{deville1993generalized}. In this approach, the classical ranking ratio method is often employed to calibrate marginal counts in the frequency table by minimizing the discrepancy between initial and newly estimated weights. The majority of methods in the synthetic reconstruction category rely on both sample and marginal data. Once weights are assigned to individual samples during fitting, they remain unchanged, making these methods deterministic \citep{fabrice2021comparing}. 

\subsection{Combinatorial optimization} 
The methods in this category aim to find an optimal solution from a finite set of objects. First, an initial synthetic population is generated, often randomly, or based on some initial heuristic. This population might not yet satisfy the required constraints such as demographic distributions, income levels, and household sizes. Next, households are drawn from the microdata to identify the best fit. Starting with randomly chosen households, the process is followed by adding, replacing, or swapping a household in the sample. If the replacement increases the fit, the household is kept \citep{templ2017simulation, voas2000evaluation}. This process is repeated until either the objective is reached or a fixed number of iterations is reached. However, the possibility of finding the optimal set can become computationally expensive if the size of the finite set is too large \citep{grotschel1995combinatorial}. Therefore, researchers have proposed heuristic algorithms to find a near-optimal solution in these scenarios, including simulated annealing \citep{templ2017simulation, huang2001comparison} and genetic algorithms \citep{katoch2021review, chen2018application, williamson1998estimation}. \citet{birkin2006synthetic} implemented a genetic algorithm to generate a synthetic population for some regions in the United Kingdom but found the model's performance to be poor as the model failed to find enough individuals from ethnic groups constituting the minority population. Similar to synthetic reconstruction, the methods in combinatorial optimization also require both the sample and marginal data and generate a synthetic population by replicating individuals.

\subsection{Statistical learning} 
The third category of methods involves simulation-based approaches \citep{fabrice2021comparing}. Unlike the other two categories, the methods in this category focus on learning the joint distribution of the variables of interest from the available microdata \citep{sun2018hierarchical, Farooq2013Simulation}. These methods avoid replication of samples by estimating a probability for different combinations, including those not present in the microdata. Markov process-based methods including the Markov Chain Monte Carlo (MCMC) simulation-based approach and the Hidden Markov Model (HMM) are a couple of widely used statistical learning-based approaches to simulate the synthetic population. The MCMC methods involve constructing the conditional distributions (e.g., income level given a set of predictors such as age and education) from microdata or zonal statistics using some parametric model (e.g., multinomial linear logistic regression). Later, the Gibbs sampler or MCMC leverages this conditional distribution of each attribute to create individuals from the joint distribution \citep{Farooq2013Simulation}. On the other hand, the HMM models the sequence of observable events that depend on internal factors or are generated by Markovian hidden state processes \citep{saadi2016hidden}. However, these studies using MCMC and HMM are limited to generating individuals and pay little attention to the hierarchical structure of households \citep{fabrice2021comparing}. \citet{casati2015synthetic} proposed an extension to the method and used a hierarchical MCMC to group individuals into households, generating a two-layered synthetic population while accounting for the household hierarchical structure. Another statistical learning-based method used by researchers to create a two-layered synthetic population is the Bayesian Network (BN). It is a probabilistic graphical model where a set of random variables (nodes) and their conditional distributions (edges) are represented in the form of a directed acyclic graph \citep{rahman2023population, young2009using}. \citet{zhang2019connected} defined a BN to consist of two main steps: (i) learning the network structure describing the dependence among related variables and (ii) estimating the parameters to learn the conditional distribution. \citet{sun2015bayesian} showed that BN can capture complex dependence and higher-order interactions within different variables by concisely abstracting the population structure. To further improve upon capturing the strong interdependencies within a household, \citet{sun2018hierarchical} proposed a multinomial hierarchical mixture model. The proposed framework uses a two-level hierarchical data structure and integrates a multilevel latent class model \citep{vermunt2003multilevel} to capture the interdependencies. The different statistical learning methods discussed above use a joint probability distribution to overcome the lack of heterogeneity which could not be resolved by synthetic reconstruction and combinatorial optimization. However, a major drawback of the statistical learning methods is that they fail to satisfy the conditional and marginal distributions simultaneously. Therefore, studies suggest using synthetic reconstruction as a post-processing step after generating a suitable representative population sample using statistical learning. For example, \citet{casati2015synthetic} and \citet{rahman2023population} used generalized ranking to post-process output from MCMC and BN, respectively.

\subsection{Deep generative methods} \label{sec:learning-based}

Recent advances in computer science techniques have allowed researchers to overcome the limitations of traditional methods with the help of Deep Generative modeling techniques. Researchers have categorized these methods as statistical learning due to their ability to learn the joint distributions \citep{fabrice2021comparing}. Unlike other statistical learning methods, however, deep generative methods do not require post-processing of the generated samples and can easily deal with many attributes. A deep learning approach involves learning a comprehensive representation from sample tables containing detailed information and using a generative neural network to synthesize a generative table. This process enables the creation of new data that aligns with the joint distribution of the sample tables. Researchers have deployed Generative Adversarial Networks (GAN), including Tabular GAN, conditional tabular GAN, and Copula GAN, to create new data to improve the disaggregated records and generate more representative and diverse datasets \citep{kotnana2022using, xu2018synthesizing, arkangil2022deep}. Moreover, \citet{lederrey2021datgan} proposed using a Directed Acyclic Tabular GAN (DATGAN) that involved integrating expert knowledge. The authors provided the neural networks with a structure of variables, which allowed them to avoid overfitting and remove possible biases. In addition to these methods, researchers also proposed the use of Variational Autoencoder (VAE) to synthesize synthetic populations \citep{borysov2019generate, borysov2021introducing}. VAE uses unsupervised learning to determine the latent variables from the training data (encoder) and use them to generate new data (decoder). \citet{borysov2019generate} found VAE to be computationally efficient while outperforming the statistical learning-based methods for higher dimensions. However, different generative models proposed by researchers focused on generating individual data and did not incorporate the joint household-individual structure for the synthetic population generated. \citet{aemmer2022generative} overcame the limitation by using a Conditional-VAE (CVAE) capable of synthesizing household and individual data simultaneously without any need for post-processing and grouping. The proposed method involved using Household CVAE to generate synthetic households and used them alongside the latent variables of the Individual CVAE decoder to enable combining individuals with households. Nonetheless, the model fails to capture the relationship between the individuals living in the same household. Moreover, using two CVAEs increases the computational demand as it involves training two models. 

Unlike the existing deep learning approach for population synthesis, the proposed approach can generate data conforming to marginal distributions outside the training data (Microdata), such as the census tract marginal distribution. Moreover, the proposed method integrates households and individuals more flexibly through microdata restructuring, eliminating the need for training multiple models. 

\section{Population Synthesis Framework}\label{sec:framework}

\begin{figure}[!ht]
    \centering
    \includegraphics[scale=0.53]{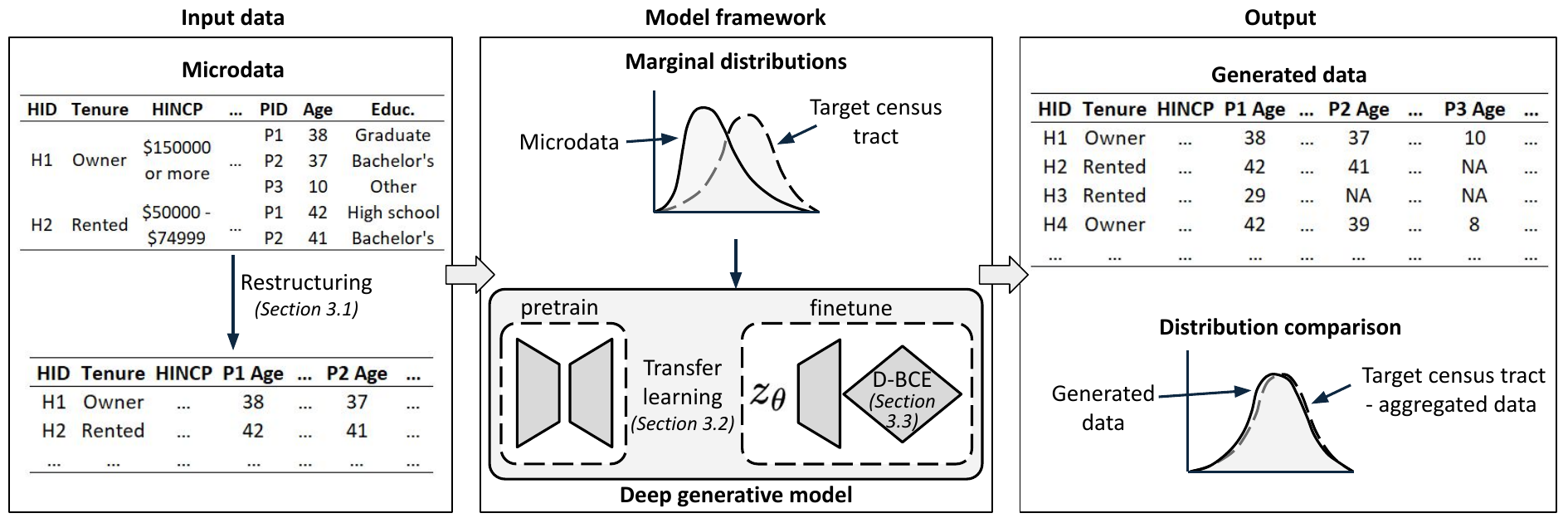}
    \caption{The end-to-end deep generative pipeline for synthetic household-individual inventory development}
    \label{fig:pipeline}
\end{figure}

In this study, we introduce a deep generative population synthesis framework, as shown in Figure \ref{fig:pipeline}, that leverages the learning of a joint distribution from microdata, ensuring the generation of synthetic households and individuals whose marginal distribution matches that of the target census tract. This framework lays the groundwork for comprehensive data generation processes with distribution shifts. Moreover, its applicability extends beyond population synthesis to other data types, especially those with divergent distributions between training and target data. The method includes three main steps: (1) restructuring the state-level microdata to facilitate the learning of joint household-individual and individual-individual associations, (2) constructing a transfer learning pipeline to allow the deep generative model to learn joint distributions in microdata and generate synthetic population conforming to distinct marginal distributions, (3) devising a decoupled binary cross-entropy loss function to enable the creation of new synthetic individuals rather than solely replicating those in the microdata. The following sections provide detailed explanations for each step.

\subsection{Microdata restructure} \label{sec:micro-restructurture}

Microdata includes details about both households and individuals. Our objective is to create a synthetic household and individual inventory that can capture (1) the connection between households and persons (i.e., household-individual) and (2) the correlation among individuals within the same household (i.e., individual-individual). To achieve this, we need to learn a high-dimensional joint distribution that captures these relationships. 

\begin{figure}[!ht]
    \centering
    \includegraphics[scale=0.7]{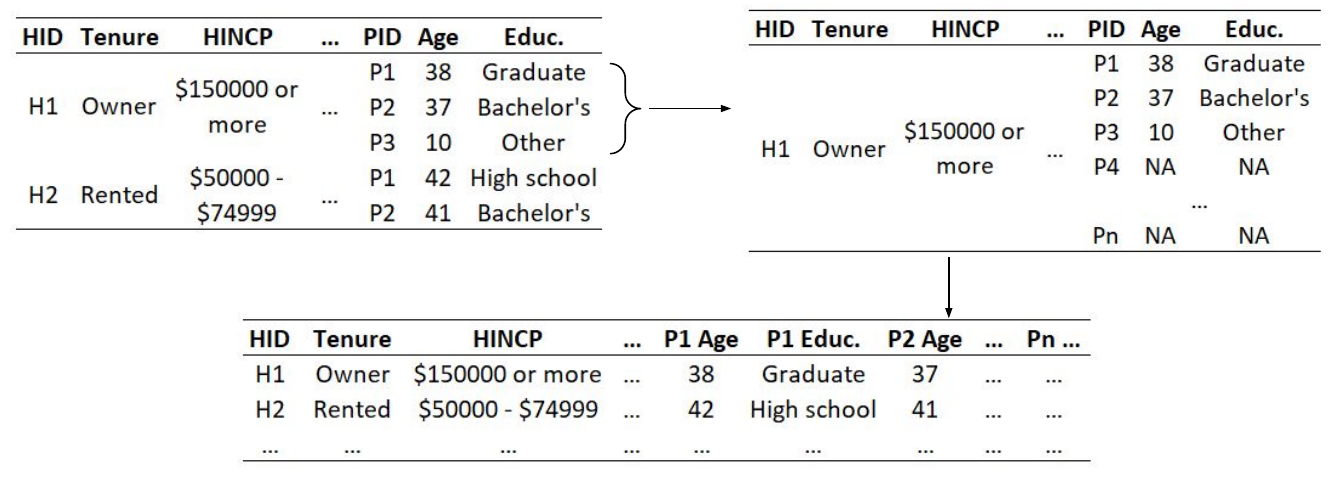}
    \caption{Data restructuring procedure illustration}
    \label{fig:data-restructure}
\end{figure}

Current methodologies commonly introduce conditional variables \citep{aemmer2022generative} or domain expertise \citep{lederrey2021datgan} into the model to capture variable relationships. This is because the data structure in their loss functions cannot represent household-individual and individual-individual relationships. Specifically, existing data structures consolidate households and individuals based on household ID, creating multiple records within one household, such as H1-P1, H1-P2, and H1-P3, as illustrated in Figure \ref{fig:data-restructure}. However, this setup leads to these records being processed separately, treating them as independent individual inputs. Consequently, we cannot effectively learn the relationship between individuals within the same household. Therefore, the traditional organization of population datasets, where one household is divided into multiple person records, hinders our ability to capture relationships between individuals within the same household and between individuals and households. 

To overcome the limitations of the existing population data structure that impact the accuracy of population synthesis, we propose a new way of restructuring microdata, wherein individuals belonging to the same household are added into the same row in the population table, such as H1-P1-P2-P3. This restructuring enables the model to effectively learn the relationships between households and individuals, as well as among individuals within the same household.

Figure \ref{fig:data-restructure} outlines the data restructuring process. First, the microdata's household table and person table are merged according to the household ID. Subsequently, we determine the maximum number of persons in the household in the dataset and set the maximum count as the window size $N_{\text{window}}$, namely, the maximum number of persons that can be present in a household according to the microdata. Then, for each household, we expand the number of corresponding persons to match $N_{\text{window}}$. If a household has fewer persons than the maximum size of $N_{\text{window}}$, the remaining person records are filled with "NA". Finally, we organize the persons within each household into a single row. Notably, during our later experiments, we find sorting persons based on features such as age and education level helps the model learn the relations between persons within the same household (i.e., individual-individual). In this way, each row in the table corresponds to a household record with all persons under it. This arrangement enables using existing record-level loss functions to learn the representation of a household.

\subsection{Parameter-efficient transfer learning under distribution shifts} \label{sec:transfer-learning}

\begin{figure}[!ht]
    \centering
    \includegraphics[scale=0.7]{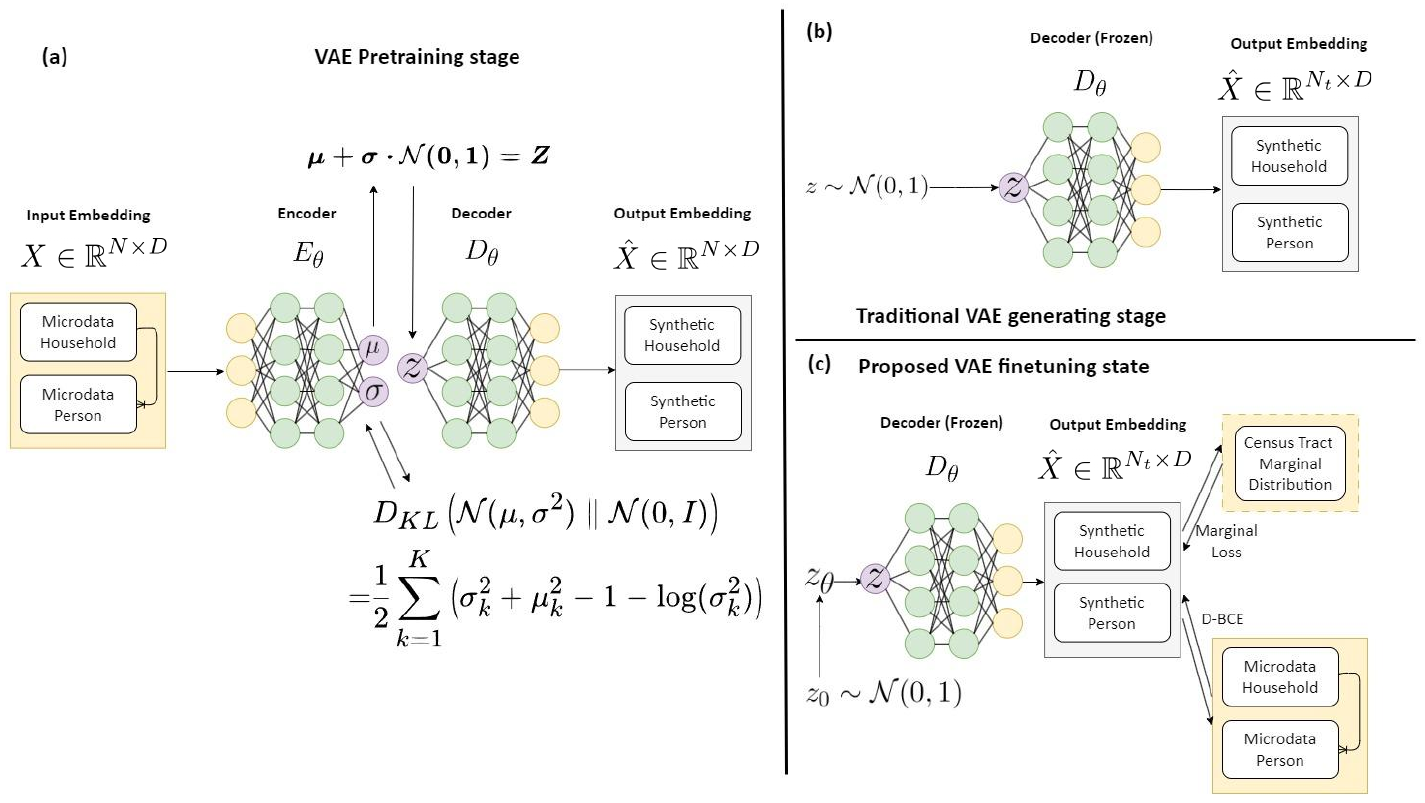}
    \caption{Parameter-efficient transfer learning procedure}
    \label{fig:transfer-learning}
\end{figure}

Deep generative methods often entail two primary steps: training and inference. Traditional methods employ a loss function to compute reconstruction errors, establishing a direct mapping between inputs and outputs. While this ensures consistency with the statistical patterns observed in microdata, it also results in the model learning the joint distribution of the microdata, as shown in Figure \ref{fig:transfer-learning}. Consequently, the inference stage generates data that tends to align with this joint distribution rather than matching the targeted marginal distribution at the census tract level.

Since the trained model can learn the joint distribution of microdata well and generate a realistic synthetic population, intuitively, we would like to retain the trained model's learning ability that withholds in the already trained parameters based on microdata and transfer it to generate data that conforms to a different distribution. This concept refers to transfer learning under distribution shifts. Transfer learning enables the generation of data that conforms to the targeted marginal distribution of specific census tracts without compromising its original capacity to produce realistic household and individual records that are consistent with the microdata. We achieve transfer learning by introducing a fine-tuning step into the traditional training and inference process (Figure \ref{fig:transfer-learning}). This approach draws inspiration from research on adversarial attacks in generative neural networks \citep{sun2021adversarial} and parameter-efficient fine-tuning techniques in large language models \citep{xu2023parameter}. 

As shown in the transfer learning procedure (Figure \ref{fig:transfer-learning}), the input to the VAE-based population synthesis pipeline is a matrix $X \in \mathbb{R}^{N \times D}$ comprises embeddings of each household and its individuals' attributes obtained from microdata, where $N$ is the number of households in microdata and $D$ is the number of household and individual attributes after one-hot encoding (see Section \ref{Sec:data}). Although the input is treated as a matrix with a customizable batch size to achieve parallel acceleration, each record is still processed independently by the network encoder $E_\theta$ and decoder $D_\theta$. This information is compressed into the latent variable $\mathbf{Z}$, a high-dimensional matrix that encapsulates all necessary details for the decoder $D_{\theta}$ to reconstruct a realistic embedding of the input \citep{chan2024tutorial}. In this stage (Figure \ref{fig:transfer-learning}(a)), the latent space $\mathbf{Z}$ is regularized to approximate a normal distribution using KL-Divergence ($D_{KL}$).

Traditionally, during the inference stage, generate tasks (Figure \ref{fig:transfer-learning}(b)) often involve sampling $\mathbf{Z}$ directly from a normal distribution and inputting it into a well-trained decoder $D_\theta$ to generate a realistic synthetic output embedding vector $\hat{X}$. However, traditional VAE can only generate data that follows the same distribution as the input data, which differs from our objective. 

In the proposed fine-tuning step (Figure \ref{fig:transfer-learning}(c)), the latent space is set as a trainable matrix $\mathbf{Z_\theta}$. Since our objective is to produce a number of households for a specific census tract, which often differs from the size of the input microdata ($N$), $\mathbf{Z_\theta}$ is sized $\mathbb{R}^{N_t \times D}$. Here, $N_t$ represents the desired number of households to generate for a target census tract, and $D$ is the number of population attributes after encoding (see Section \ref{Sec:data}). The decoder $D_\theta$ still processes each row of $\mathbf{Z_\theta}$ independently, either individually or in a batch. After obtaining the synthetic embedding vector $\hat{X_i}$, we then organize the vectors into a matrix $\hat{X}\in \mathbb{R}^{N_t \times D}$. The Root Mean Square Error (RMSE) loss is calculated using the marginal distribution of the census tract from the census table. This loss information is backpropagated through the frozen $D_\theta$ to the trainable matrix $\mathbf{Z_\theta}$. The $\mathbf{Z_\theta}$ is periodically updated until the output $\hat{X} \in \mathbb{R}^{N_t \times D}$ closely matches the target marginal distribution. This fine-tuning process is parameter-efficient because only the input latent variables are updated, while the model's parameters remain fixed.

\begin{figure}[!ht]
    \centering
    \includegraphics[scale=0.7]{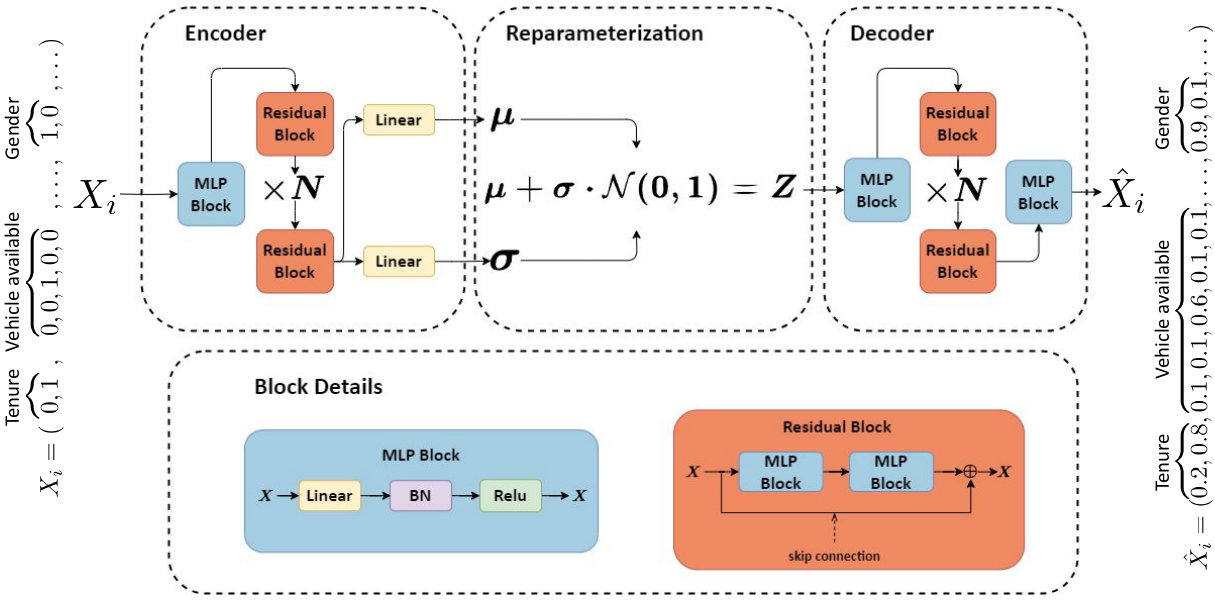}
    \caption{VAE structure in the proposed deep generative framework}
    \label{fig:VAE}
\end{figure}

The proposed transfer learning procedure can be applied to various generative models, including Variational Autoencoder (VAE), Generative Adversarial Networks (GAN), and Diffusion Models. To demonstrate the effectiveness of the transfer learning approach, we use the VAE model in this study (Figure \ref{fig:VAE}). An autoencoder (AE) comprises two components: an encoder and a decoder. The encoder compresses data from a higher-dimensional space into a lower-dimensional space, known as the latent space, while the decoder reconstructs the latent space back into the higher-dimensional space. Both components are trained together using a loss function that aims to reconstruct the input accurately at the output. We harness the capabilities of an autoencoder to learn continuous representations of the microdata's heterogeneous features within the latent space \citep{suh2023autodiff}. In contrast, a Variational Autoencoder (VAE) introduces a constraint on the latent distribution, forcing it to follow a normal distribution. This ensures that the latent variable is smooth and continuous, thereby enabling the latent space with generative capabilities.

Each row of the input is a vector ($X_i$), representing a restructured household record that includes both household and individual characteristics after one-hot encoding (see Section \ref{Sec:data}). The outputs of the encoder are two vectors, representing the mean $\mu$ and log variance $\log \sigma$ of the latent space. These values are then reparameterized to derive the latent space variable $\mathbf{Z}$, which serves as the input for the decoder. The reparameterization process is expressed as $Z = \mu + \epsilon \odot \log\sigma$, where $\epsilon \sim \mathcal{N}(0, 1)$. The decoder outputs the probability distribution for each variable. For instance, for the variable "tenure", which is the first term of $\hat{X_i} = (\overbrace{0.2, 0.8}^{\text{Tenure}},\overbrace{0.1, 0.1, 0.6, 0.1, 0.1}^{\text{Vehicle available}}, \dots, \overbrace{0.9, 0.1}^{\text{Gender}}, \dots)$, the possible outcomes are "Owned" or "Rented". The decoder produces two probabilities, such as [0.2, 0.8], corresponding to the likelihood of "Owned" and "Rented".

The encoder includes six feedforward neural networks. In the last layer, separate fully connected layers and Batch Normalization (BN) are used to output $\mu$ and $\log \sigma$. The decoder also includes six feedforward neural networks, with the last layer's output dimension set to $D$ (i.e., the number of population attributes after encoding). After applying one-hot encoding to a set of vectors for the same variable, softmax is used to generate the probability of each variable. Each feedforward neural network in both the encoder and decoder consists of a fully connected layer, followed by a BN layer and a ReLU (Rectified Linear Unit) activation layer.

\subsection{Decoupled binary cross-entropy (D-BCE)} \label{sec:D-BCE}

\begin{figure}[!ht]
    \centering
    \includegraphics[scale=0.6]{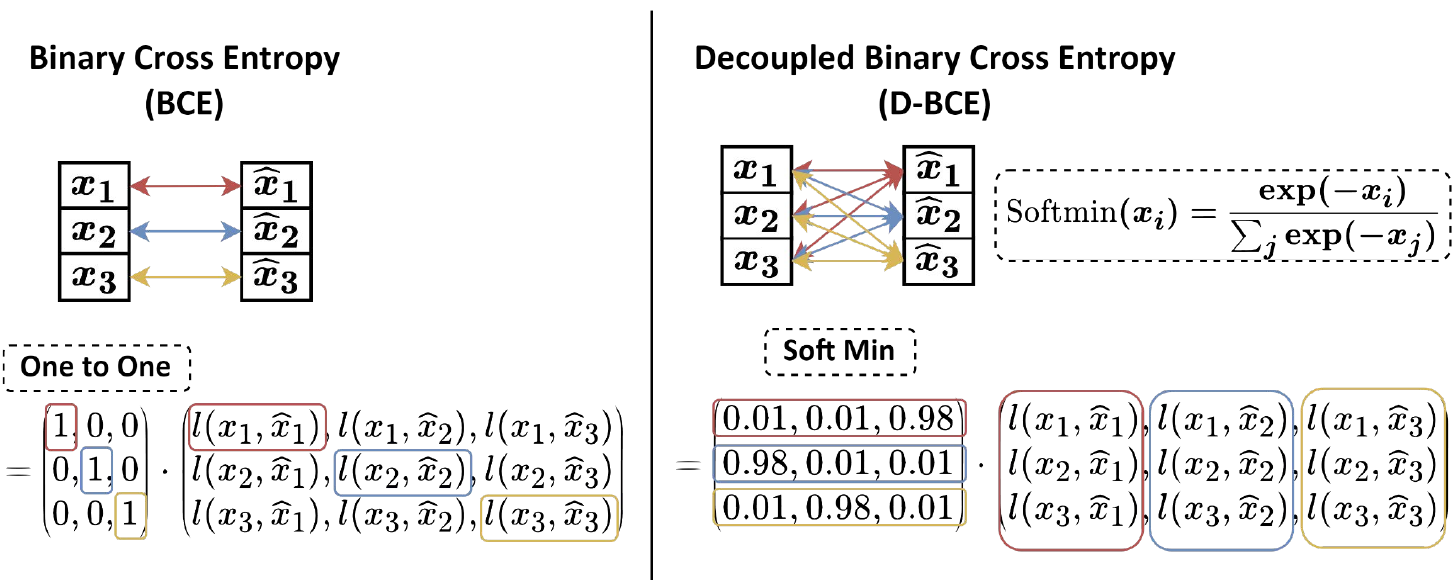}
    \caption{Illustrative comparison between BCE and D-BCE}
    \label{fig:D-BCE}
\end{figure}

During the training of the generative model to emulate the microdata, the Binary Cross-Entropy (BCE) loss function is commonly utilized, represented mathematically as follows:
\begin{equation}
    \text{BCE loss: } l(x_{i,j}, \hat{x}_{i,j}) = -\frac{1}{N}\sum_{i=1}^{N} \sum_{j=1}^{D} \left[ \hat{x}_{i,j} \log x_{i,j} + (1-\hat{x}_{i,j}) \log (1-x_{i,j}) \right]
    \label{eq:bce}
\end{equation}
where $N$ represents the number of households in microdata, $D$ denotes the number of population variables after onehot encoding, $x_{i,j}$ represents the actual value of the $j$-th variable in the $i$-th record, and $\hat{x}_{i,j}$ is the generated value of the $j$-th variable in the $i$-th record. The BCE loss function necessitates a one-to-one match between the generated and microdata household at the record level, as illustrated in Figure \ref{fig:D-BCE}. However, this also results in the generated household's marginal distribution matching that of the microdata, contradicting our objective of producing authentic households that adhere to the target marginal distribution at each census tract level. 

We propose that each generated household does not need to precisely match its corresponding input data record. Instead, as long as it closely resembles any record in the microdata, we consider it an authentic generated instance. Guided by this principle, we have formulated the Decoupled Binary Cross-Entropy (D-BCE) loss function. The detailed procedure is illustrated in Algorithm \ref{alg:decoupled_bce}.

For row $i$ in $\hat{X}$, $\hat{x}_i$, we compute the D-BCE loss with row $j$ in $X$, $x_j$, resulting in a vector $\mathsf{bce}_i$ of length $N$ (Algorithm \ref{alg:decoupled_bce}, Line 5). Subsequently, for each $\mathsf{bce}_i$ calculated, we compute its softmin, yielding a vector $\mathsf{softIndex}_i$ of length $N$. The next step involves taking the inner product of $\mathsf{softIndex}_i$ and $\mathsf{bce}_i$, resulting in the soft minimum loss $\mathsf{softmin}_i$ (Algorithm \ref{alg:decoupled_bce}, Lines 7-8). Using the minimum value of $\mathsf{bce}_i$ directly would lead to a non-smooth gradient. Drawing insights from knowledge distillation \citep{hinton2015distilling}, we convert the hard label to a soft label to obtain a smooth gradient and enable the computation of the label's gradient. The soft label also makes it possible to compute the label's gradient. Finally, we average all record-specific soft minimum losses $\mathsf{softmin}_i$ to derive the Decoupled Binary Cross-Entropy loss (Algorithm \ref{alg:decoupled_bce}, Line 11). This process is illustrated in the Figure \ref{fig:D-BCE}.

Because we relaxed the loss calculation from a strict one-to-one correspondence to resembling any record in the microdata, one potential concern with the D-BCE arises, namely, the generated data might lean towards being similar to only a few records in the microdata, potentially impacting the diversity of the generated data. To address this, we propose the D-BCE Norm KL (Kullback–Leibler), which quantifies the diversity of the generated data. This involves summing all the soft minimum losses $\mathsf{softmin}_i$ to obtain the vector $\mathsf{softIndex}$ representing the entire generated data. We then calculate the KL divergence between $\mathsf{softIndex}$ and a uniform distribution, resulting in the D-BCE Norm KL (Algorithm \ref{alg:decoupled_bce}, Line 12).  

\begin{algorithm}
    \caption{Decoupled Binary Cross-Entropy (D-BCE)}
    \begin{algorithmic}[1]
    \Require Micro data table $X \in \mathbb{R}^{N \times D}$, Generated data table $\hat{X} \in \mathbb{R}^{N_t \times D}$
    \Ensure Decoupled Binary Cross-Entropy Loss, Decoupled Binary Cross-Entropy Norm KL
    
    \State Initialize vector $\mathsf{softIndex}$ of length $N$ to zeros
    \For{each row $i$ in $\hat{X}$ ($\hat{x}_i$)}
        \State Initialize vector $\mathsf{bce}_i$ of length $N$
        \For{each row $j$ in $X$ ($x_j$)}
            \State $\mathsf{bce}_i[j] \gets \mathsf{BCE}(\hat{x}_i, x_j)$ \Comment{Compute Binary Cross-Entropy Loss}
        \EndFor
        \State $\mathsf{softIndex}_i \gets \mathsf{softmin}(\mathsf{bce}_i)$ \Comment{Compute soft minimum index of $\mathsf{bce}_i$}
        \State $\mathsf{softmin}_i \gets \langle \mathsf{softIndex}_i, \mathsf{bce}_i \rangle$ \Comment{Compute the soft minimum loss}
        \State $\mathsf{softIndex} \gets \mathsf{softIndex} + \mathsf{softIndex}_i$ \Comment{Accumulate softIndex}
    \EndFor

    \State $\text{Decoupled BCE Loss} \gets \frac{1}{N_t} \sum_{i=1}^{N_t} \mathsf{softmin}_i$ \Comment{Average the soft minimum losses}
    \State $\text{Decoupled BCE Norm KL} \gets \text{KL}(\text{Uniform}(N), \mathsf{softIndex})$ \Comment{KL divergence}

    \State \Return $\text{Decoupled BCE Loss}$, $\text{Decoupled BCE Norm KL}$
    \end{algorithmic}
    \label{alg:decoupled_bce}
\end{algorithm}

The D-BCE has two advantages over traditional BCE in population synthesis. First, its relaxed loss calculation allows the joint distribution of different variables in the generated household to match a household record in the microdata, while permitting differences in the marginal distribution of the generated and original households. Second, the proposed D-BCE accommodates microdata with varying data lengths as input ($N_t \neq N$), facilitating accurate modeling of real data distribution. 

Similar to the original BCE, D-BCE also involves modeling the joint distribution of generated synthetic households and microdata. A higher D-BCE indicates that the generated synthetic household is not similar to the microdata. Conversely, a lower D-BCE value suggests overfitting of the generated synthetic household to the microdata, potentially leading to a lack of diversity in the generated synthetic households. It's important to note that an appropriate D-BCE value should remain within the same order of magnitude as it was at the end of the pretraining phase. This shows that fine-tuning does not compromise the model's ability to produce authentic synthetic household data. Additionally, the mathematical formulation of D-BCE Norm KL inherently helps prevent overfitting the microdata, thereby preserving diversity in the generated synthetic household data.

\section{Experimental Design}\label{sec:design}

\subsection{Data}\label{Sec:data}

This study utilized ACS PUMS data (microdata) \citep{PUMS2022} and ACS Census Data Tables (census tables) \citep{ACS2022} for both training and testing purposes, focusing on the data from the year 2021. We opted for ACS data over AHS due to its broader coverage and granularity. AHS is limited to data from approximately 100,000 housing units across only 35 metro areas and selected states, whereas ACS encompasses around 3.5 million addresses annually and offers information at national, state, and county levels, down to the tract and block group levels \citep{ACSAHS2023}. Consequently, developing a synthetic population based on ACS data enhances the generalizability of our findings to wider geographic regions. 

As shown in Table \ref{tab:inventory-attributes}, for households, we included the following variables: TEN (Tenure), HINCP (Household Income), R18 (Presence of Persons Under 18 Years in the Household), R65 (Presence of Persons 65 Years and Over in the Household), HHL (Household Language), and VEH (Vehicles Available). For individual persons, the variables considered were AGEP (Age), SEX (Sex), and SCHL (Educational Attainment). This selection of variables aims to showcase the performance of the proposed model by encompassing various types of data. Specifically, we intentionally selected household attributes such as R18 and R65, as well as the individual's age, to assess the model's performance, as detailed in Section \ref{sec:result}. Depending on the intended application of this synthetic inventory, the list can be expanded accordingly.

\begin{table}[!ht]
\centering
\caption{Household and individual attributes}
\label{tab:inventory-attributes}
\resizebox{\textwidth}{!}{%
\begin{tabular}{ll|ll}
\hline
\multicolumn{2}{c|}{\textbf{Household}}                                                                                                                                                                                                                                                                          & \multicolumn{2}{c}{\textbf{Individual}}                                                                                                                                                                                                                                                                                                                                        \\ \hline
\multicolumn{1}{l|}{Variable} & \multicolumn{1}{c|}{Description}                                                                                                                                                                                                                                        & \multicolumn{1}{l|}{Variable}              & \multicolumn{1}{c}{Description}                                                                                                                                                                                                                                                                                          \\ \hline
\multicolumn{1}{l|}{TEN}      & Tenure (Owned, Rented)                                                                                                                                                                                                                                   & \multicolumn{1}{l|}{SEX}                   & Sex (Male, Female)                                                                                                                                                                                                                                                                                                      \\ \hline
\multicolumn{1}{l|}{VEH}      & \begin{tabular}[c]{@{}l@{}}Vehicles Available (No vehicle available, 1 vehicle, \\ 2 vehicles, 3 vehicles, 4 or more vehicles)\end{tabular}                                                                                                                             & \multicolumn{1}{l|}{\multirow{3}{*}{AGEP}} & \multirow{3}{*}{\begin{tabular}[c]{@{}l@{}}Age (Under 5, 5-9, 10-14, 15-19,\\ 20-24, 25-29, 30-34, 35-39, \\ 40-44, 45-49, 50-54, 55-59, \\ 60-64, 65-69, 70-74, 75-79, \\ 80-84, 85 and over) \end{tabular}} \\ \cline{1-2}
\multicolumn{1}{l|}{R18}      & \begin{tabular}[c]{@{}l@{}}Presence of Persons Under 18 Years in the Household \\ (Yes, No)\end{tabular}                                                                                                                                                        & \multicolumn{1}{l|}{}                      &                                                                                                                                                                                                                                                                                                                          \\ \cline{1-2}
\multicolumn{1}{l|}{R65}      & \begin{tabular}[c]{@{}l@{}}Presence of Persons 65 Years and Over in the Household \\ (Yes, No)\end{tabular}                                                                                                                                                     & \multicolumn{1}{l|}{}                      &                                                                                                                                                                                                                                                                                                                          \\ \hline
\multicolumn{1}{l|}{HHL}      & \begin{tabular}[c]{@{}l@{}}Household Language (English only, Spanish, \\ Other Indo-European languages)\end{tabular}                                                                                                                                                    & \multicolumn{1}{l|}{\multirow{2}{*}{SCHL}} & \multirow{2}{*}{\begin{tabular}[c]{@{}l@{}}Educational Attainment (NA, Less than high \\ school graduate, High school graduate (or \\ equivalency), Some college or associate's \\  degree, Bachelor's degree, Graduate or \\ professional degree)\end{tabular}}                                                    \\ \cline{1-2}
\multicolumn{1}{l|}{HINCP}    & \begin{tabular}[c]{@{}l@{}}Household Income (Less than \$5,000, \$5,000 to \$9,999,  \\ \$10,000 to \$14,999, \$15,000 to \$19,999, \$25,000 to \$34,999, \\ \$35,000 to \$49,999, \$50,000 to \$74,999, \$75,000 to \$99,999, \\ \$100,000 to \$149,999, \$150,000 or more)\end{tabular} & \multicolumn{1}{l|}{}                      &                                                                                                                                                                                                                                                                                                                          \\ \hline
\end{tabular}%
}
\end{table}

In this paper, continuous variables such as income are transformed into categorical variables for both data and methodological reasons. First, the target marginal distribution for each census tract is presented in a categorical format. To ensure alignment between the synthetic population's marginal distribution and the target distribution, it is necessary to convert numeric values to categorical variables for consistency. Second, previous research in deep learning-based tabular data generation has demonstrated that using categorical variables instead of numerical values can improve generation accuracy \citep{borysov2019generate}. 

\begin{figure}[!ht]
    \centering
    \includegraphics[scale=0.6]{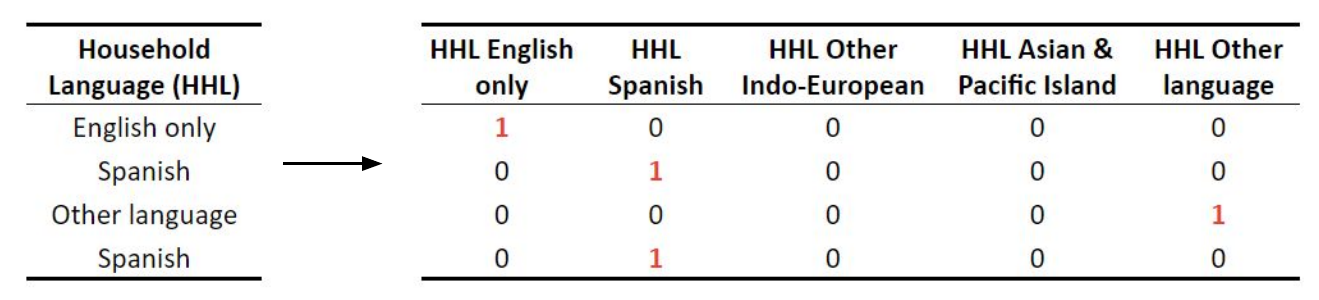}
    \caption{An illustrative example of one-hot encoding}
    \label{fig:onehot}
\end{figure}

Following the data type conversion, we apply the one-hot encoding to transform categorical variables into a machine learning-compatible format \citep{seger2018investigation}. This method involves converting each categorical value into a new binary feature, where a value of 1 indicates the presence of the category, and 0 means its absence. Figure \ref{fig:onehot} illustrates the one-hot embedding process for the attribute of household language.

We selected Delaware as our study site and North Carolina as the transferability test site. The microdata for Delaware consists of $N=18,641$ household samples, while for North Carolina, it comprises 198,037 household samples. Following data restructuring and attribute one-hot encoding, we applied the proposed method to a randomly selected census tract in Delaware. 

\subsection{Model setup}

One-hot encoding ensures the input data is in a consistent format for pre-training, but it also results in sparse feature representations, with a higher number of 0s than 1s, as illustrated in Figure \ref{fig:onehot}. This imbalance poses challenges to traditional BCE, particularly in reconstructing the minority class during pre-training. Our preliminary experiments also confirmed this, showing that traditional BCE made the model hard to converge. This challenge can be addressed by adopting the focal loss (FL) technique \citep{lin2017focal}, as described in Eq. (\ref{eq:focal-loss}). FL enhances traditional BCE by incorporating a modulation factor that reduces the loss assigned to well-classified examples, enabling the model to focus more on difficult-to-classify samples.

\begin{equation}
      \text{FL}(x_{i,j}, \hat{x}_{i,j}) = -\frac{1}{N}\sum_{i=1}^{N} \sum_{j=1}^{D} \left[ \alpha \hat{x}_{i,j} (1 - x_{i,j})^\gamma \log x_{i,j} + (1 - \alpha) (1-\hat{x}_{i,j}) x_{i,j}^\gamma \log (1-x_{i,j}) \right]
      \label{eq:focal-loss}
\end{equation}
where $\alpha$ is a weighting parameter ranging from 0 to 1, used to balance positive and negative samples. Its value is determined by the ratio of 0s to 1s in the dataset. $\gamma$ controls the influence of the modulation factor, with a larger $\gamma$ making the model focus more on difficult-to-classify samples, and vice versa. When $\gamma=0$, the focal loss is equivalent to traditional BCE. The other parameters remain the same as in Eq. (\ref{eq:bce}). Focal loss has proven effective in facilitating model pre-training during our experiments. Therefore, we use the FL for pre-training and D-BCE for fine-tuning in this study. 

We employ the Lion optimizer \citep{chen2023symbolic} with an initial learning rate of 0.001, a batch size equal to the dataset size, and training lasting for 4000 epochs. Starting from the 1000th epoch, the learning rate is exponentially decayed, reaching a minimum of 0.0001. A machine with the following specifications is used in our experiments: GPU: NVIDIA GeForce RTX 4060 Ti with 8GB RAM. CPU: 12th Gen Intel(R) Core(TM) i7-12700F.

\subsection{Performance metrics}
\label{sec:performance-metrics}
To evaluate the performance of the proposed models, we employed two statistical metrics, including root mean square error (RMSE) and KL divergence. 

Root Mean Squared Error (RMSE) measures the differences between predicted and actual values. Lower RMSE values indicate better model performance.

\begin{equation}
    \text{RMSE} = \sqrt{\frac{1}{n} \sum_{i=1}^{n} (x_i - \hat{x}_i)^2}
\end{equation}
where $n$ is the number of data points, $x_i$ is the actual value, and $\hat{x}_i$ is the predicted value. When comparing two distributions, the RMSE quantifies how closely the synthetic data matches the original data by computing the average squared difference between corresponding percentages in the two distributions and then taking the square root of that average.

Kullback-Leibler (KL) divergence measures how one probability distribution diverges from the second, expected probability distribution. KL divergence is often used comparatively; the model with the lower KL divergence is considered to be a better approximation of the true distribution. A KL divergence score of zero indicates that the two distributions are identical. 

\begin{equation}
    D_{KL}(\mathbf{\hat{X}} || \mathbf{X}) = \sum_{i} (\hat{x}_i + \epsilon) \log\left(\frac{\hat{x}_i + \epsilon}{x_i + \epsilon}\right)
\end{equation}
where $\mathbf{\hat{X}}$ is the distribution of the synthetic population, and $\mathbf{X}$ is the ground truth marginal distribution. $x_i$ is the $i^{th}$ element of ground truth marginal distribution $\mathbf{X}$ and $\hat{X}_i$ is the $i^{th}$ element of synthetic population marginal distribution $\mathbf{\hat{X}}$. $\epsilon$ is a small positive value, often representing an error or tolerance. 

\section{Results}\label{sec:result}

\subsection{Realism of synthetic population using pre-trained model}\label{sec:pretrain-delaware}

The pre-trained model aims to accurately capture statistical relationships among households and individuals, as well as interactions between individuals. We utilize the pre-trained VAE model to generate a synthetic population that is the same size as the microdata. The realism of the synthetic population is assessed by how closely the distributions derived from the microdata match the distribution of the synthetic data produced by the pre-trained model. This comparison is made for both individual attributes (e.g., household income, household language) and joint variables (e.g., household income-household language). 

\begin{figure}[!ht]
    \centering
    \includegraphics[scale=0.7]{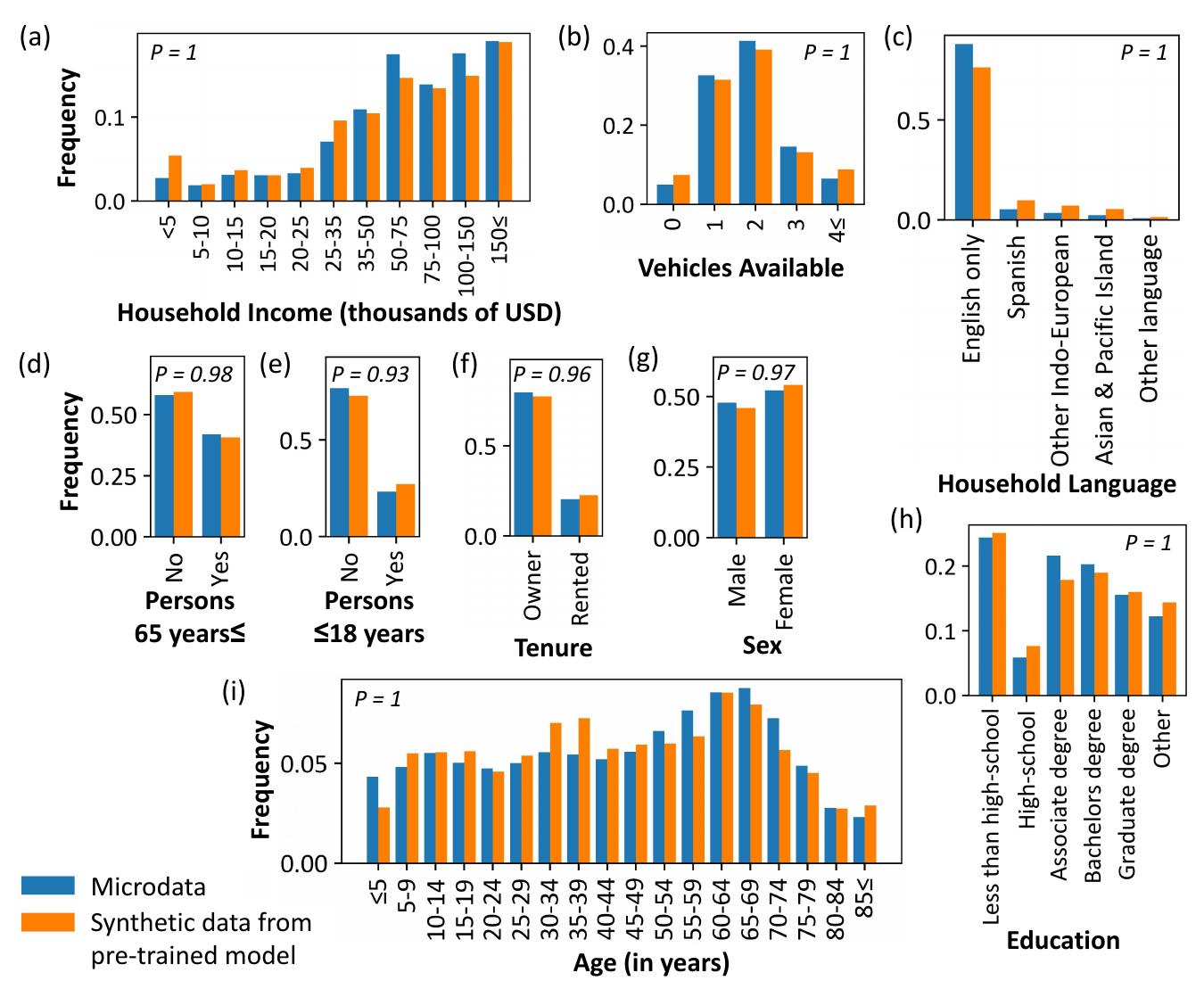}
    \caption{Attribute distribution comparison between microdata and pre-trained VAE. (a)-(f) Household attributes, (g)-(i) Individual attributes. }
    \label{fig:pretrain-results}
\end{figure}

We use both household and individual attributes to demonstrate the performance of the pre-trained model, as shown in Figure \ref{fig:pretrain-results}. The bar plot illustrates a strong resemblance in marginal distributions between the microdata and synthetic data generated by the pre-trained model, indicating that the pre-trained model can effectively generate realistic synthetic household data that aligns well with the microdata. We further conducted a chi-square test to compare the two distributions (null hypothesis). Across all household and individual attributes, we obtained a \textit{p}-value ($P$) greater than 0.9, suggesting that the null hypothesis is not rejected and there is no evidence to suggest a statistically significant difference between the marginal distribution from the pre-trained model and that of the microdata.

\begin{table}[!ht]
\caption{Performance of pre-trained model on individual attributes}
\label{tab:pretrain-stats-individual}
\small
\begin{adjustbox}{width=\textwidth,center}
\begin{tabular}{@{}l|*{6}{p{1.5cm}}|*{3}{p{1.5cm}}|*{1}{p{1.5cm}}@{}}
\hline
\multicolumn{1}{c|}{}   & 
\multicolumn{6}{c|}{\textbf{Household}}& \multicolumn{3}{c|}{\textbf{Individual}} & \multicolumn{1}{c}{}  \\ 
\hline
\textbf{} 
& 
\begin{tabular}{@{}l@{}}\textbf{Tenure} \\ \textbf{(TEN)} \end{tabular}
& 
    \begin{tabular}{@{}l@{}}\textbf{Vehicles}\\ \textbf{Available} \\ \textbf{(VEH)}\end{tabular}
&
    \begin{tabular}{@{}l@{}}\textbf{Household}\\ \textbf{Language} \\ \textbf{(HHL)}\end{tabular}
&
    \begin{tabular}{@{}l@{}}\textbf{Household}\\ \textbf{Income} \\ \textbf{(HINCP)}\end{tabular}
&
\begin{tabular}{@{}l@{}}\textbf{Persons}\\ \textbf{$\leq 18$-yrs} \\ \textbf{(R18)}\end{tabular}
&
\begin{tabular}{@{}l@{}}\textbf{Persons}\\ \textbf{$\geq$ $65$-yrs} \\ \textbf{(R65)}\end{tabular}
& \begin{tabular}{@{}l@{}}\textbf{Age} \\ \textbf{(AGEP)}\end{tabular}
& \begin{tabular}{@{}l@{}}\textbf{Education} \\ \textbf{(SCHL)}\end{tabular}
& \begin{tabular}{@{}l@{}}\textbf{Sex} \\ \textbf{(SEX)}\end{tabular}
& \textbf{Mean} \\ 
\hline
\begin{tabular}{@{}l@{}}\textbf{- RMSE}\end{tabular}& 0.0210 & 0.0198 & 0.0598 & 0.0164 & 0.0388 & 0.0129 & 0.0091 & 0.0200 & 0.0190 & 0.0241 \\
\hline
\begin{tabular}{@{}l@{}}\textbf{- KL}\end{tabular} & 0.0013 & 0.0095 & 0.0442 & 0.0177 & 0.0039 & 0.0003 & 0.0138 & 0.0081 & 0.0007 & 0.0111 \\
\hline
\end{tabular}
\end{adjustbox}
\end{table}

We utilize the metrics listed in Section \ref{sec:performance-metrics} to evaluate the performance of the pre-trained model. The results of these metrics are presented in Table \ref{tab:pretrain-stats-individual}. The "-" sign suggests that a smaller value (closer to 0) implies better performance. The results show that the pre-trained model achieves a KL divergence score near zero, indicating that the individual attributes of the synthetic population closely approximate those in the microdata. 

\begin{figure}[!ht]
    \centering
    \includegraphics[scale=0.75]{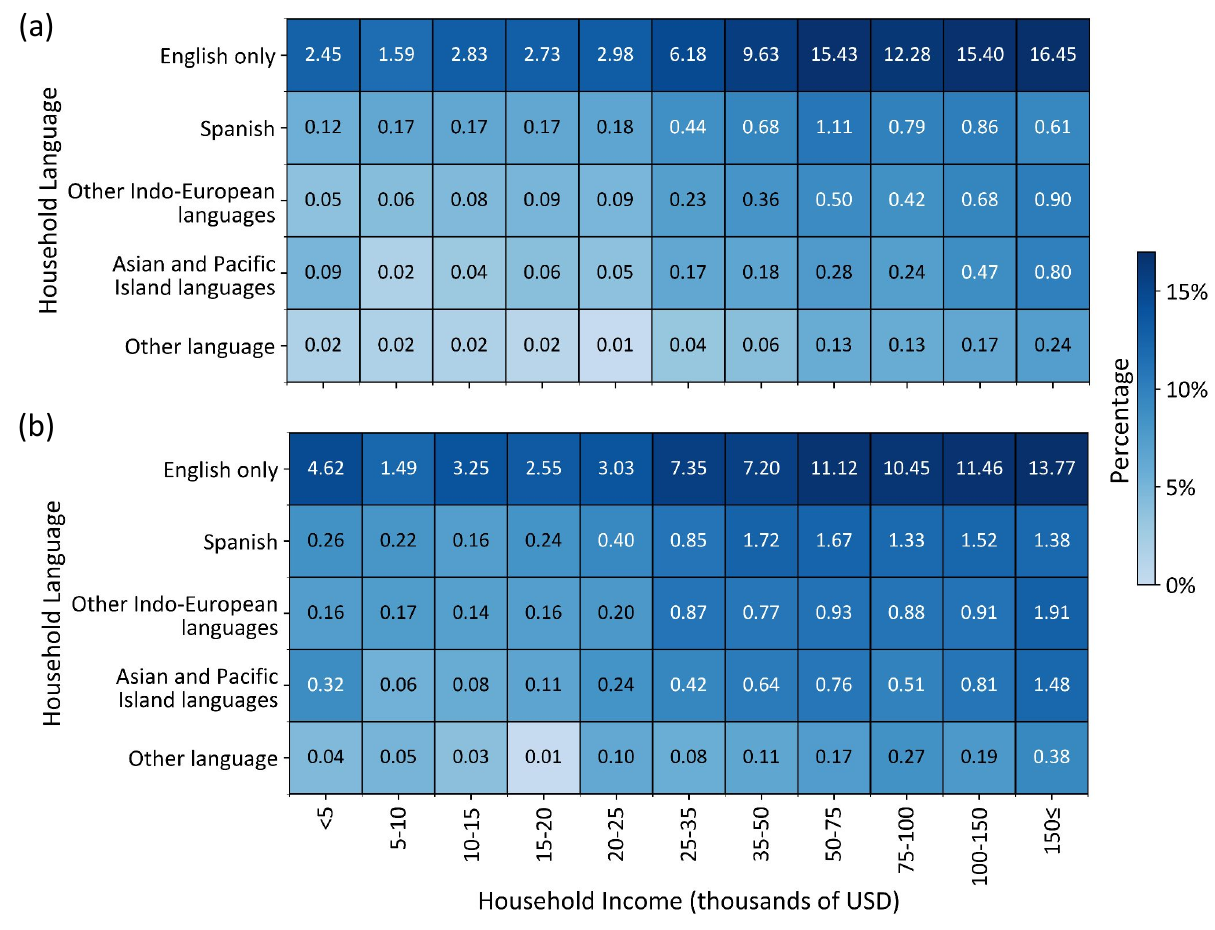}
    \caption{Joint distribution between different synthetic population attributes. (a) Microdata (b) Pretrain.}
    \label{fig:pretrain-attribute-relationship}
\end{figure}

Additionally, we analyze the correlations between various attributes within a household to validate that the pre-trained VAE can accurately capture statistical relationships among household attributes. Figure \ref{fig:pretrain-attribute-relationship}(a) illustrates the relationship between household language and household income in the microdata, while Figure \ref{fig:pretrain-attribute-relationship}(b) displays the same relationship in the synthetic household data generated by the pre-trained VAE. Each box is colored based on the log-scaled value of the percentage. The log transformation is applied to highlight differences among small values and to moderate extremely large values. Without this transformation, the English-only rows would dominate the coloring scheme, making it difficult to visualize differences in the other categories. The close resemblance between the two colormaps further confirms the pre-trained VAE's ability to produce realistic synthetic household data. We further conducted a chi-square test to compare the joint distribution of all 36 pairs of variables from the synthetic population with those in the microdata. The results yield a $p$-value greater than 0.99 for all comparisons, suggesting that the null hypothesis is not rejected, and there is no evidence to indicate a statistically significant difference between the two joint distributions.

Similarly, we evaluated the performance of the pre-trained model by calculating the RMSE and KL divergence between the joint distribution of microdata and the synthetic population across all 36 pairs of variables. Table \ref{tab:pretrain-stats-joint} shows consistently low RMSE and KL divergence scores, indicating that the joint distributions of the synthetic population closely approximate those in the microdata. For example, the RMSE for the joint distribution of age and household income between the microdata and the synthetic population is 0.0021. This indicates that the synthetic population's distribution for this pair of variables deviates from the microdata by an average error of 0.21\%. While KL divergence has been employed to assess synthetic population performance, these evaluations often focus on specific household sizes and selected variable pairs \citep{zhang2019connected}. Since our joint household-individual population development is highly unique, there are no available benchmarks for comparing KL divergence. Nevertheless, the low KL divergence value in Table \ref{tab:pretrain-stats-joint}(b) indicates that the joint variable distributions of the synthetic population closely match those in the microdata.

\begin{table}[h]
\caption{Performance of pre-trained model on joint attributes. (a) RMSE, (b) KL Divergence.}
\label{tab:pretrain-stats-joint}
\small
\centering
\begin{tabular}{cccccccccc}
\hline
     & TEN    & VEH    & HHL    & HINCP  & R18    & R65    & AGEP   & SCHL   & SEX    \\
\hline
TEN  & ------ &        &        &        &        &        &        &        &        \\
VEH  & 0.0161 & ------ &        &        &        &        &        &        &  (a)   \\
HHL  & 0.0419 & 0.0210 & ------ &        &        &        &        &        &        \\
HINCP& 0.0141 & 0.0078 & 0.0107 & ------ &        &        &        &        &        \\
R18  & 0.0348 & 0.0228 & 0.0484 & 0.0136 & ------ &        &        &        &        \\
R65  & 0.0180 & 0.0166 & 0.0303 & 0.0102 & 0.0250 & ------ &        &        &        \\
AGEP & 0.0067 & 0.0041 & 0.0049 & 0.0021 & 0.0076 & 0.0088 & ------ &        &        \\
SCHL & 0.0139 & 0.0086 & 0.0125 & 0.0052 & 0.0178 & 0.0121 & 0.0037 & ------ &        \\
SEX  & 0.0154 & 0.0166 & 0.0236 & 0.0102 & 0.0116 & 0.0203 & 0.0053 & 0.0119 & ------ \\
\hline
\end{tabular}

\vspace{3mm}
\begin{tabular}{cccccccccc}
\hline
     & TEN    & VEH    & HHL    & HINCP  & R18    & R65    & AGEP   & SCHL   & SEX    \\
\hline
TEN  & ------ &        &        &        &        &        &        &        &        \\
VEH  & 0.0200 & ------ &        &        &        &        &        &        &  (b)   \\
HHL  & 0.0494 & 0.0604 & ------ &        &        &        &        &        &        \\
HINCP& 0.0403 & 0.0747 & 0.0684 & ------ &        &        &        &        &        \\
R18  & 0.0071 & 0.0247 & 0.0595 & 0.0337 & ------ &        &        &        &        \\
R65  & 0.0041 & 0.0202 & 0.0478 & 0.0275 & 0.0126 & ------ &        &        &        \\
AGEP & 0.0226 & 0.0538 & 0.0516 & 0.0660 & 0.0458 & 0.0715 & ------ &        &        \\
SCHL & 0.0119 & 0.0305 & 0.0422 & 0.0525 & 0.0189 & 0.0146 & 0.1224 & ------ &        \\
SEX  & 0.0019 & 0.0157 & 0.0252 & 0.0258 & 0.0014 & 0.0028 & 0.0188 & 0.0102 & ------ \\
\hline
\end{tabular}
\end{table}

According to the performance metrics summarized here, it is evident that the pre-trained model effectively captures relationships between different variables, generating synthetic data with minimal deviations from the microdata. These findings affirm the capability of the proposed VAE model during the pre-training stage to produce realistic synthetic household and individual records, achieving features 1, 2, \& 3 of a realistic synthetic population. Therefore, we are assured of using this pre-trained model to produce synthetic household and individual data at the census tract level.

\subsection{Synthetic population of census tracts using fine-tuned model} \label{sec:finetune-delaware}

\begin{figure}[!ht]
    \centering
    \includegraphics[scale=0.6]{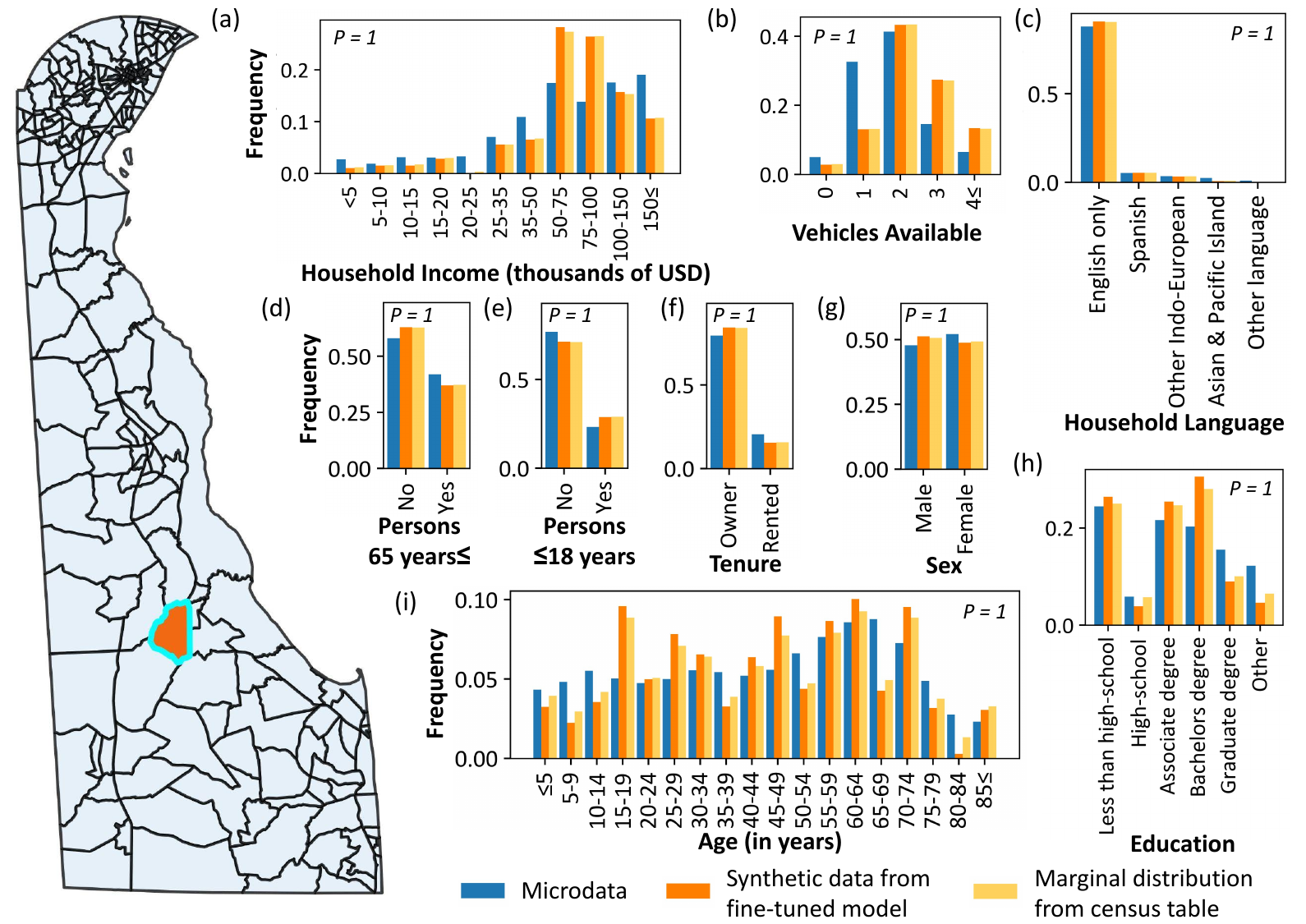}
    \caption{Distributions of generated household and individual attributes using the finetuned model for a randomly selected census tract in Delaware. (a)-(f) household attributes, (g)-(i) individual attributes.}
    \label{fig:fine-tuned-DE-model-results}
\end{figure}

The objective of the fine-tuning step in the proposed deep generative pipeline is to shift the distribution that the synthetic population adheres to from the microdata distribution to the target marginal distribution at the census tract level. Figure \ref{fig:fine-tuned-DE-model-results} shows the final results of the fine-tuned model. It is evident that attributes in the synthetic household-individual inventory (illustrated by the orange bar) significantly departed from those in the microdata (represented by the blue bar), while closely approximating the target marginal distribution (displayed by the yellow bar) at the census tract level. The chi-square test yielded a \textit{p}-value of 1 for all attributes, indicating that the generated synthetic household-individual data from the fine-tuned model aligns accurately with the marginal distribution provided by the census table. This underscores the accuracy of the synthetic population and the effectiveness of the proposed transfer learning approach under the distribution shift.

We further evaluate the performance of the generated synthetic household-individual inventory using the six statistical metrics listed in Section \ref{sec:performance-metrics}. To establish a comparison baseline, we utilize the marginal distribution observed in microdata. Microdata serves as our baseline because existing deep learning methods typically concentrate on generating synthetic individuals that conform to the marginal distribution observed in microdata. Consequently, the baseline performance is assessed by comparing the microdata's marginal distribution with the target marginal distribution from the census table. As shown in Table \ref{tab:finetune-stats}, we demonstrate the enhancements offered by our proposed method in achieving a more realistic characterization of household-individual characteristics at the census tract level. For example, looking at Figure \ref{fig:fine-tuned-DE-model-results}, existing deep learning-based synthetic populations tend to overestimate the number of households with incomes over 150k and underestimate those with incomes 75-100k because they rely on microdata for generation. In each category, we can see that our synthetic data substantially outperforms the baseline methods by aligning the synthetic data more accurately with the marginal distribution provided in the census table, enabling more precise estimations of these numbers. The close approximation of distributions at the census tract level indicates that our synthetic population is geographically realistic, fulfilling feature 4 of a realistic synthetic population. 

\begin{table}[!ht]
\caption{Population synthesis performance of fine-tuned model}
\label{tab:finetune-stats}
\small
\begin{adjustbox}{width=\textwidth,center}
\begin{tabular}{@{}l*{1}{p{1.5cm}}|*{6}{p{1.5cm}}|*{3}{p{1.5cm}}|*{1}{p{1.5cm}}@{}}
\hline
\multicolumn{2}{c|}{}   & 
\multicolumn{6}{c|}{\textbf{Household}}& \multicolumn{3}{c|}{\textbf{Individual}} & \multicolumn{1}{c}{}  \\ 
\hline
\textbf{} & \textbf{} 
& 
\textbf{Tenure} 
& 
    \begin{tabular}{@{}l@{}}\textbf{Vehicle}\\ \textbf{Available}\end{tabular}
&
    \begin{tabular}{@{}l@{}}\textbf{HH.}\\ \textbf{Language}\end{tabular}
&
    \begin{tabular}{@{}l@{}}\textbf{HH.}\\ \textbf{Income}\end{tabular}
&
\begin{tabular}{@{}l@{}}\textbf{Persons}\\ \textbf{$\leq 18$-yrs}\end{tabular}
&
\begin{tabular}{@{}l@{}}\textbf{Persons}\\ \textbf{$\geq$ $65$-yrs}\end{tabular}
& \textbf{Age} 
& \textbf{Edu.} 
& \textbf{Sex} 
& \textbf{Mean} \\ 
\hline
\multirow{2}{*}{\textbf{- RMSE}} & \textbf{Baseline} & 0.0468 & 0.1088 & 0.0142 & 0.0574 & 0.0579 & 0.0471 & 0.0180 & 0.0469 & 0.0285 & 0.0473 \\
& \textbf{Syn. HI.} & 0.0021 & 0.0018 & 0.0011 & 0.0034 & 0.0024 & 0.0021 & 0.0068 & 0.0169 & 0.0057 & 0.0047 \\
\hline
\multirow{2}{*}{\textbf{- KL}} & \textbf{Baseline} & 0.0072 & 0.1508 & 0.0167 & 0.1338 & 0.0089 & 0.0046 & 0.0520 & 0.0425 & 0.0016 & 0.0465 \\
& \textbf{Syn. HI.} & 0.0000 & 0.0001 & 0.0000 & 0.0575 & 0.0000 & 0.0000 & 0.0166 & 0.0093 & 0.0001 & 0.0093 \\
\hline
\end{tabular}
\end{adjustbox}
\end{table}

\subsection{Transferability of the proposed deep generative population synthesis framework} \label{sec:application-in-NC}

\begin{figure}[!ht]
    \centering
    \includegraphics[scale=0.6]{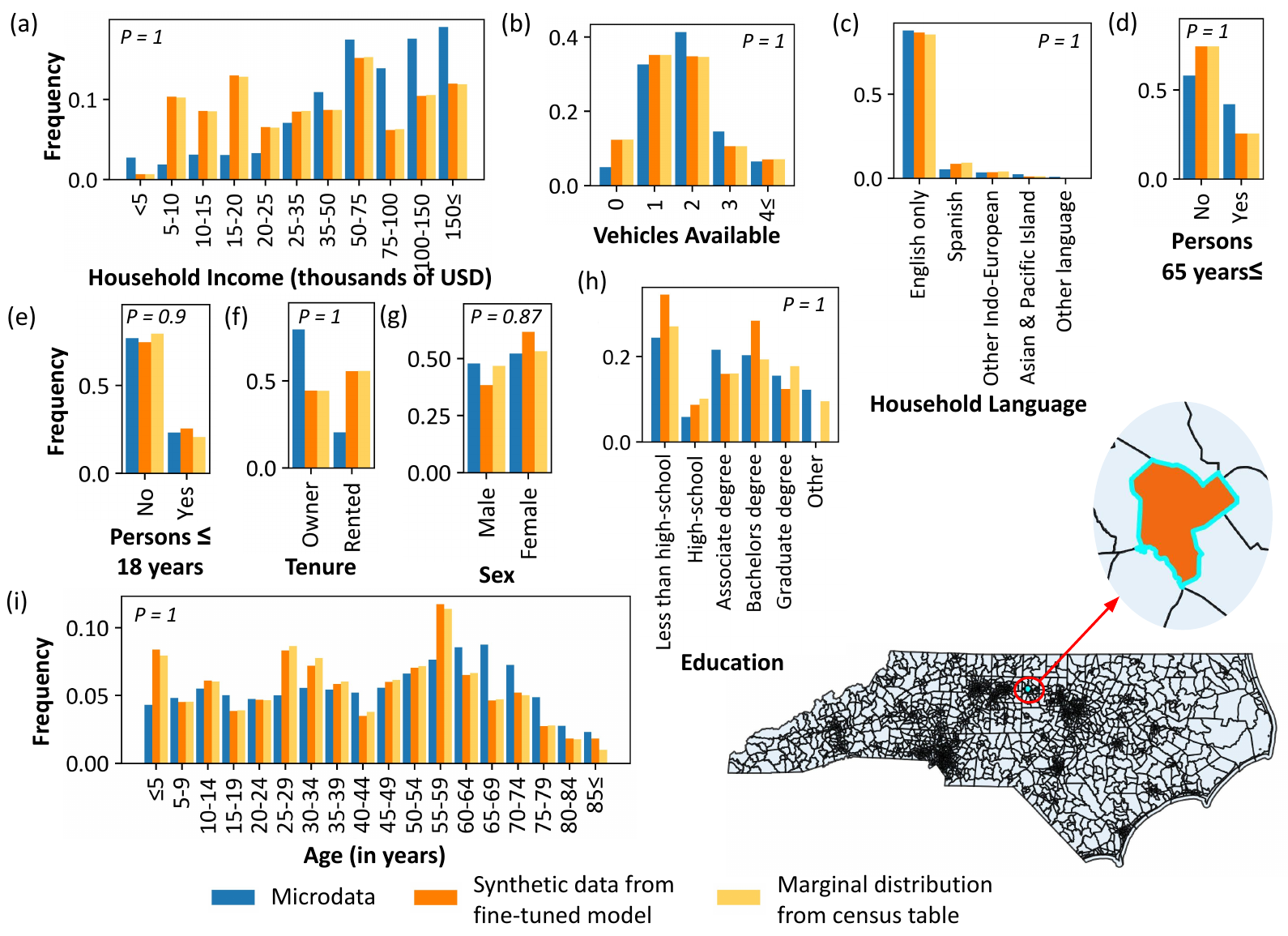}
    \caption{Distributions of generated household and individual attributes using the finetuned model for a randomly selected census tract in North Carolina. (a)-(f) household attributes, (g)-(i) individual attributes.}
    \label{fig:fine-tuned-NC-model-results}
\end{figure}

Ideally, we aim for the proposed deep generative framework for population synthesis to have the flexibility to be applied across various locations. To assess its transferability, we tested the framework in North Carolina. With 198,037 households in the microdata for North Carolina, we utilized it for training purposes and subsequently generated a household-individual population dataset for a census tract in North Carolina. The results depicted in Figure \ref{fig:fine-tuned-NC-model-results} exhibit a similar pattern to those presented in Figure \ref{fig:fine-tuned-DE-model-results}. Notably, the marginal distribution of the synthetic population (represented by the orange bar) differs significantly from that of the microdata (depicted by the blue bar), while closely matching the target marginal distribution at the census tract level (indicated by the yellow bar). The chi-square test conducted between the target marginal distribution from the census table and the synthetic population yields \textit{p}-values of 1 for most variables, except for R18 (persons below 18 years of age) (\textit{p}-value = 0.9) and Sex (\textit{p}-value = 0.87). This highlights the robust transferability of the proposed framework to other regions.

\subsection{Population synthesis privacy}

A key concern in data synthesis is safeguarding the privacy of the training dataset used to generate the synthetic data, aiming to prevent the disclosure of sensitive information, especially pertaining to human subjects. This becomes even more crucial when researchers intend to create a synthetic population using privately collected data. Fortunately, in this study, the household and individual data within the microdata released by the US Census Bureau have already been anonymized to uphold data privacy. Therefore, data privacy is not the primary concern in this specific case. Nevertheless, we anticipate that our population synthesis pipeline can be widely applicable in various scenarios, particularly in cases when the marginal distribution of the training data differs from the targeted marginal distribution of the synthetic data. In this context, we aim to ensure that the transfer learning procedure, particularly the fine-tuning step, does not compromise the privacy-preserving capability of the pre-trained model.

\begin{figure}[!ht]
    \centering
    \includegraphics[scale=0.5]{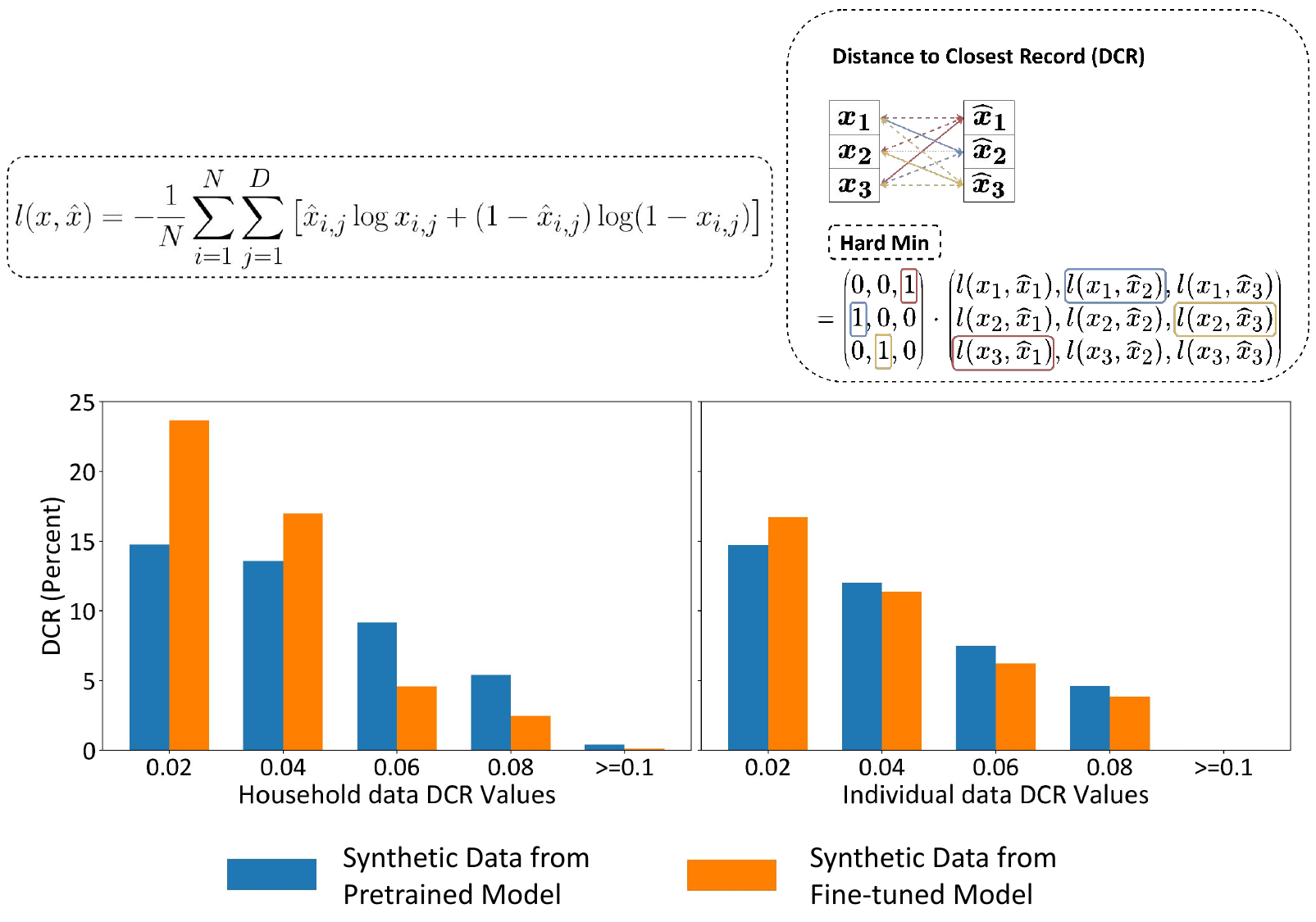}
    \caption{Population synthesis privacy assessment}
    \label{fig:synthesis-privacy}
\end{figure}

We assess privacy using Distances to Closest Records (DCR). DCR identifies the record in the microdata that the generated synthetic record most closely resembles (i.e., has the least distance) and calculates the distance between records using simple BCE. Our objective is to ensure there is no statistically significant difference (at alpha=0.5) between the results of the fine-tuned model and the pre-trained model. As shown in Figure \ref{fig:synthesis-privacy}, the fine-tuning step maintains a similar level of privacy as the pre-trained model. Furthermore, we grouped the DCR values into different bins and performed the Kolmogorov-Smirnov (K-S) test to compare the differences between distributions of households' and individuals' DCR. The resulting p-values for the household and individual data were $0.87$ and $1.00$, respectively, both exceeding the threshold of $0.05$. This indicates no statistically significant difference between the DCR distributions of the pre-trained and fine-tuned models. That is to say, transferring learning effectively preserves privacy and does not worsen the privacy performance of the generative model. However, it is important to note that VAE is not the leading method in terms of privacy preservation capabilities compared to other generative methods such as AutoDiff \citep{suh2023autodiff}. In scenarios involving the use of private household survey data for synthetic population generation or employing the proposed pipeline in other sensitive data generation tasks, the generative module, VAE in this paper, can be replaced with other methods to meet privacy requirements.

\section{Discussion}\label{sec:discussion}

The proposed deep generative population synthesis framework enables the generation of records that extend beyond the microdata. While we aim to capture the joint distribution between attributes and closely align the aggregated distribution with the marginal distribution, there still exist some discrepancies. Consequently, unrealistic synthetic records that deviate from reality may be generated. However, manually identifying these "unrealistic" records is laborious and challenging. 

To assess the consistencies between each household and the individuals it includes in the generated synthetic population (Feature 2), we specifically kept some variables in both household and individual attributes in our test. For example, we included the presence of individuals under 18 years (R18) or 65 years and above (R65) in the household, as well as the individual's age. Intuitively, an individual's age $\geq$ 65 would correspond to the household attributes R65 being flagged as 1. Otherwise, this record would indicate faulty generated records. Applying this criterion, we conducted a sanity check of the generated synthetic household-individual inventory. In the census tract highlighted in Figure \ref{fig:fine-tuned-DE-model-results}, we identified 158 households (11\%) with contradictory attributes (R65 flagged in the household, but lacking individuals aged $\geq$ 65 years) in synthetic household-individual data, indicating inconsistent records. This suggests that, although we can achieve decent realism at the distribution level by achieving low RMSE and KL divergence of both individual attributes and joint variables (Figure \ref{fig:pretrain-results} and \ref{fig:pretrain-attribute-relationship}, Table \ref{tab:pretrain-stats-individual} and \ref{tab:pretrain-stats-joint}), record-level realism requires extra attention. This limitation is inherent in all deep generative methods, yet it is seldom discussed and reported in existing studies. Prior research often prioritizes distribution accuracy over record-level correctness \citep{sun2015bayesian, saadi2016hidden, borysov2019generate, aemmer2022generative}. However, the accuracy of household records is crucial for subsequent tasks such as equity assessment and household decision-making analysis. We intentionally retained these records as they serve as both an indicator of the framework's advantage (i.e., generating data beyond microdata) and its limitation (i.e., producing faulty records). The record-level examination does not detract from the methodological contributions presented in this paper. Instead, it highlights a direction for future population synthesis research to enhance. To further enhance the proposed framework and minimize detectable unrealistic records, we can leverage human expertise during the generation phase. For instance, we can employ human-in-the-loop learning techniques \citep{cao2023reinforcement}, to identify and label rule-violating records. Subsequently, the learning algorithm can be informed to avoid generating such records, thus reducing the rate of unrealistic records.

\begin{table}[!ht]
\caption{Unrealistic samples with inconsistent household character and individual attributes}
\begin{adjustbox}{width=\textwidth,center}
\begin{tabular}{cc|ccccccccc}
\hline
\textbf{IID}            & \textbf{HHID}        & \textbf{Tenure}          & \textbf{\begin{tabular}[c]{@{}c@{}}Vehicles\\ Available\end{tabular}} & \textbf{\begin{tabular}[c]{@{}c@{}}Household\\ Language\end{tabular}} & \textbf{\begin{tabular}[c]{@{}c@{}}Household Income\\ (thousands of USD)\end{tabular}} & \textbf{\begin{tabular}[c]{@{}c@{}}Persons\\ $\leq$ 18years\end{tabular}} & \textbf{\begin{tabular}[c]{@{}c@{}}Persons\\ $\geq$ 65years \end{tabular}} & \textbf{\begin{tabular}[c]{@{}c@{}}Age\\ (in years)\end{tabular}} & \textbf{Educational} & \textbf{Sex}                \\ \hline
29                      &                      &                          &                                                                       &                                                                       &                                                                                        &                                                                           &                                                                           & 25 to 29                                                          & college or associate & Male                        \\
30                      & \multirow{-2}{*}{10} & \multirow{-2}{*}{Renter} & \multirow{-2}{*}{2}                                                   & \multirow{-2}{*}{English only}                                        & \multirow{-2}{*}{\$75,000 to \$99,999}                                                 & \multirow{-2}{*}{no}                                                      & \multirow{-2}{*}{no}                                                      & 30 to 34                                                          & High school          & Female                      \\ \hline
\multicolumn{1}{|c}{31} &                      &                          &                                                                       &                                                                       &                                                                                        &                                                                           & \cellcolor[HTML]{EA9999}                                                  & \cellcolor[HTML]{EA9999}40 to 44                                  & college or associate & \multicolumn{1}{c|}{Female} \\
\multicolumn{1}{|c}{32} & \multirow{-2}{*}{11} & \multirow{-2}{*}{Owner}  & \multirow{-2}{*}{3}                                                   & \multirow{-2}{*}{English only}                                        & \multirow{-2}{*}{\$50,000 to \$74,999}                                                 & \multirow{-2}{*}{no}                                                      & \multirow{-2}{*}{\cellcolor[HTML]{EA9999}yes}                             & \cellcolor[HTML]{EA9999}45 to 49                                  & High school          & \multicolumn{1}{c|}{Male}   \\ \hline
33                      &                      &                          &                                                                       &                                                                       &                                                                                        &                                                                           &                                                                           & 15 to 19                                                          & High school          & Male                        \\
34                      &                      &                          &                                                                       &                                                                       &                                                                                        &                                                                           &                                                                           & 50 to 54                                                          & Bachelor             & Female                      \\
35                      & \multirow{-3}{*}{12} & \multirow{-3}{*}{Owner}  & \multirow{-3}{*}{4 or more}                                           & \multirow{-3}{*}{English only}                                        & \multirow{-3}{*}{\$100,000 to \$149,999}                                               & \multirow{-3}{*}{no}                                                      & \multirow{-3}{*}{no}                                                      & 55 to 59                                                          & High school          & Male                      \\
\hline
\end{tabular}
\end{adjustbox}
\label{tab:sanity-check}
\end{table}

Furthermore, in this paper, only nine attributes are included to evaluate the proposed framework. However, after recategorizing each variable, applying one-hot embedding, and restructuring the data with fifteen individuals per record, each entry already spans a window size of 462 columns. In practice, we may need to incorporate more household and individual attributes to better characterize household decision-making or equity assessment. Increasing the number of attributes will significantly grow the size of the training data, which requires high-performance computing resources and poses a challenge to the generative power of the deep learning model that is beyond what VAE can offer. Therefore, in future research, we plan to explore more efficient data restructuring methods while still capturing household-individual and individual-individual relationships. Additionally, we will consider replacing VAE with more powerful generative models such as transformers \citep{solatorio2023realtabformer}. A stronger generative backbone method not only enables the handling of large volumes of data and attributes but also improves the realism of synthetic records.

\section{Conclusion}\label{sec:conclusion}

This paper introduces a deep generative framework for synthetic population development. It leverages microdata, which consists of anonymized samples of real households and individuals at the state level, along with marginal distributions (representing the distribution of each attribute at the census tract level) to create a diverse inventory of households with individuals embedded. We aim to ensure this synthetic population is realistic in two main aspects: (1) the marginal distribution of household characteristics (e.g., income, tenure.) and individual attributes (e.g., age, education) aligns with the target marginal distribution provided by ACS census data tables at the census tract level; (2) the relationships between household characteristics and individual attributes, as well as correlations between individuals, accurately reflect those described in the microdata.

We employ the Variational Autoencoder (VAE) for synthetic population generation (Figure \ref{fig:VAE}). It allows for the generation of out-of-sample records, enhancing population diversity, as opposed to simply weighing and cloning microdata samples. This deep generative framework presents three methodological contributions aimed at addressing corresponding challenges in existing synthetic populations. Firstly, we introduce a data restructuring scheme (Figure \ref{fig:data-restructure}) that captures not only relationships between household and individual attributes but also relationships among individuals within a household. This approach overcomes the limitation of the existing two-step process, where individuals are first generated and then grouped into households, thus failing to capture household-individual relationships. Secondly, we propose a parameter-efficient transfer-learning approach (Figure \ref{fig:transfer-learning}) consisting of pre-training and fine-tuning. The pre-training step learns the joint distribution of household and individual characteristics in microdata, while the fine-tuning step generates households and individuals that fit a different distribution at the census tract level. This is in contrast to existing population generation that follows the same marginal distribution as the microdata, which is not realistic given the significant differences between marginal distribution microdata at the state level and marginal distribution at the census tract level (as seen in Figure \ref{fig:data-challenge}). Thirdly, we introduce a new loss function, Decoupled Binary Cross Entropy (D-BCE) (Figure \ref{fig:D-BCE}), which focuses on generating households and individuals similar to any record in the microdata, rather than strictly mirroring a specific record. This decoupling procedure relaxes one-to-one correspondences to one-to-many correspondences, enabling the aforementioned transfer learning process. 

We tested the framework in Delaware using six household attributes and three individual attributes (Table \ref{tab:inventory-attributes}). The synthetic inventory yields promising results. The pre-trained model successfully captures relationships between households and individuals, as well as among individuals. Notably, the pre-trained VAE demonstrates strong performance across all employed metrics, ensuring the realism of the generated data in subsequent steps (Figure \ref{fig:pretrain-results} and \ref{fig:pretrain-attribute-relationship}, Table \ref{tab:pretrain-stats-individual} and \ref{tab:pretrain-stats-joint}). Moreover, the results from the fine-tuned model indicate our ability to generate a synthetic household-individual inventory that aligns with the marginal distribution of various attributes at the census tract level, as provided by ACS census data tables (Figure \ref{fig:fine-tuned-DE-model-results}). Additionally, we demonstrate that our proposed model outperforms existing deep-generated inventories (Table \ref{tab:finetune-stats}). To ensure the applicability of our framework to other regions, we tested it in North Carolina (Figure \ref{fig:fine-tuned-NC-model-results}), obtaining similarly promising results, thereby confirming the transferability of our methods. Lastly, recognizing the potential adoption of our method by studies dealing with sensitive human-subject information, we examined the privacy implications of our framework using Distance to Closest Record (DCR) (Figure \ref{fig:synthesis-privacy}). The Kolmogorov-Smirnov (K-S) test indicates no statistically significant difference, affirming the privacy-preserving capability of our approach.

Future research will continue enhancing the realism, privacy protection, and generative capabilities of population synthesis. This will involve exploring more robust and privacy-preserving deep generative backbone methods, while also incorporating a wider range of household and individual attributes.

\section*{Acknowledgments}
The authors would like to acknowledge funding support from the National Science Foundation \#2209190. Any opinions, conclusions, and recommendations expressed in this research are those of the authors and do not necessarily reflect the view of the funding agencies. The authors would also like to thank the editor and the anonymous reviewers for their constructive comments and valuable insights to improve the quality of the article. 

\bibliographystyle{plainnat}

\label{sec:refs}

\end{document}